\documentclass[journal]{IEEEtran}
\usepackage{amssymb, amsmath, amsthm, graphicx, enumerate}
\usepackage{setspace}
\usepackage{algorithm}
\usepackage{algpseudocode}
\usepackage{array}
\usepackage{subfigure}
\usepackage{float}
\usepackage{amstext}
\usepackage{amsmath}
\usepackage{graphicx}
\usepackage{booktabs}
\usepackage{tabularx}
\usepackage{multirow}
\usepackage{cite}
\usepackage{amssymb}
\usepackage{color}
\usepackage{amsmath}
\usepackage{lineno}
\usepackage{amsmath}
\usepackage{lineno}
\usepackage{latexsym}
\usepackage{textcomp}
\usepackage{lipsum}
\usepackage{booktabs}
\usepackage{hyperref}

\begin{document}


\title{Modified Diversity of Class Probability Estimation Co-training for Hyperspectral Image Classification}
\author{~Yan~Ju, Lingling~Li,~\IEEEmembership{Member,~IEEE}, Licheng~Jiao,~\IEEEmembership{Fellow,~IEEE}, Zhongle~Ren,~\IEEEmembership{Student~Member,~IEEE}, Biao~Hou,~\IEEEmembership{Member,~IEEE}, Shuyuan~Yang,~\IEEEmembership{Senior~Member,~IEEE}
    \thanks{This work was supported in part by the Major Research Plan of the National Natural Science Foundation of China (No. 91438201 and No. 91438103), the Fund for Foreign Scholars in University Research and Teaching Programs (the 111 Project) (No. B07048), the Fundamental Research Funds for the Central Universities(No. XJS17108) and the China Postdoctoral Fund (No. 2017M613081)}
	\thanks{Y. Ju, L. Li, L. Jiao, Z. Ren, B. Hou and S. Yang are with the Key Laboratory of Intelligent Perception and Image Understanding of Ministry of Education of China, Xidian University, Xi'an, Shaanxi Province, 710071, China (Email: yan.juchn@gmail.com).
}
}

\maketitle
\pagestyle{empty}  
\thispagestyle{empty} 

\begin{abstract}
Due to the limited amount and imbalanced classes of labeled training data, the conventional supervised learning can not ensure the discrimination of the learned feature for hyperspectral image (HSI) classification. In this paper, we propose a modified diversity of class probability estimation (MDCPE) with two deep neural networks to learn spectral-spatial feature for HSI classification. In co-training phase, recurrent neural network (RNN) and convolutional neural network (CNN) are utilized as two learners to extract features from labeled and unlabeled data. Based on the extracted features, MDCPE selects most credible samples to update initial labeled data by combining k-means clustering with the traditional diversity of class probability estimation (DCPE) co-training. In this way, MDCPE can keep new labeled data class-balanced and extract discriminative features for both the minority and majority classes. During testing process, classification results are acquired by co-decision of the two learners. Experimental results demonstrate that the proposed semi-supervised co-training method can make full use of unlabeled information to enhance generality of the learners and achieve favorable accuracies on all three widely used data sets: Salinas, Pavia University and Pavia Center.
\end{abstract}

\IEEEpeerreviewmaketitle

\section{Introduction} 
\IEEEPARstart{W}{ith} the development of satellite imaging technique, HSI with high resolution in spectral as well as spatial is available. Observed by thousands of continuous electromagnetic spectrums, each pixel of HSI contains abundant terrain information. Hence, automatic interpretation of HSI has great potential in urban development\cite{gu2016nonlinear,li2016discontinuity}, change detection of landscape\cite{wu2014slow,lyu2016learning}, scene interpretation\cite{hu2015transferring} and regulation of natural resource\cite{olmanson2013airborne,moran1997opportunities}, etc. HSI classification is an essential task and lays the ground for the applications mentioned above.

Numerous traditional methods are proposed for HSI classification during past years, such as decision trees\cite{delalieux2012heathland}, random forest\cite{ham2005investigation}, support vector machine (SVM)\cite{melgani2004classification,gualtieri2000support}, etc. Nevertheless, these approaches utilize spectral information but ignore abundant spatial information\cite{chen2014deep}, which has been proved to be useful in improving the interpretation of HSI data\cite{plaza2009incorporation}. Thus, more and more methods based on spectral-spatial feature are presented and improve accuracy significantly\cite{bioucas2013hyperspectral,fauvel2008spectral,li2013spectral,liu2013spatial} for HSI classification.

With the increase of spectral channels and spatial variability of spectral signature, the shallow-layer classifiers mentioned above can not ensure the validity of the learned feature. As the most prosperous machine learning method nowadays, deep learning (DL) are able to extract more discriminative feature and achieve better performance than traditional shallower classifiers for HSI classification\cite{chen2014deep,mou2017deep,chen2015spectral}. Mou \emph{et al.}\cite{mou2017deep} proposed a novel RNN with a new activation function and modified gated recurrent unit (GRU) for HSI classification. This model takes the spectra as a sequential signal for the first time and acquires good performance only based on spectral feature. Spectral-spatial feature is proved to be effective in HSI classification, which promotes the study of classification models based on deep spectral-spatial feature of HSI \cite{chen2015spectral,zhao2016spectral,chen2016deep,zhong2018spectral}. For example, Zhao \emph{et al.}\cite{zhao2016spectral} combined the CNN-based spatial feature and the BLDE-based spectral feature into image interpretation, and proposed a spectral-spatial feature-based classification (SSFC) method, which achieved good classification performance. A 3-D CNN-based feature extraction model is also proposed in\cite{chen2016deep} in order to extract discriminative spectral-spatial feature of HSI and the classification results also show fully using of spectral-spatial information can improve accuracy considerably.

The deep models mentioned above are supposed to be trained with large number of labeled data in a supervised way. As the depth of the deep model increases, more and more parameters need to be trained and more labeled data is demanded to prevent overfitting. While in practical classification issues, labeled data is much less than unlabeled data, as manual labeling of remote sensing data is costly in finance and manpower. Furthermore, utilizing costly labeled data while neglecting accessible unlabeled data is a great waste. In such case, semi-supervised learning (SSL) is proposed to enlarge the labeled data by recovering unlabeled samples.

There are many classic methods for SSL, such as generative model with EM\cite{dempster1977maximum}, self-training\cite{yarowsky1995unsupervised}, co-training\cite{blum1998combining}, transductive SVM\cite{bie2004convex}, graph based method\cite{blum2001learning}, etc.
Recently, plenty of semi-supervised models have been proposed for HSI classification\cite{dopido2013semisupervised,zhang2014modified,samiappan2015semi,camps2007semi}. In\cite{zhang2014modified}, a modified co-training method for HSI classification is presented. This method extracts the spectral feature and 2-D Gabor feature from spatial domain in order to co-train two classifiers from distinct views. Shallow feature is extracted to handle complex classification problems in these semi-supervised models, which is not as robust and discriminative as deep feature is. Hence, Ratle \emph{et al.}\cite{ratle2010semisupervised} combined deep learning with semi-supervised learning and proposed a semi-supervised neural networks (SSNNs) to deal with HSI classification problems. By adding a regularizer to the loss function for training neural networks, this method promotes classification accuracy considerably. However, without studying the class balance, it fails to represent the distributive characteristics of HSI data and provide poor accuracies across all classes\cite{he2009learning}.

In conclusion, there are two main difficulties in fulfilling accurate HSI classification\cite{camps2005kernel,he2009learning}: 1) classifiers are prone to be overfitting and can not extract discriminative feature because of limited labeled HSI data with many spectral channels and great spatial variability; 2) classifiers tend to provide a severely imbalanced accuracies across classes, with the majority class having high accuracy and the minority class having abysmal accuracy if labeled data is class-imbalanced.

To solve these problems, a semi-supervised method is proposed based on MDCPE co-training of RNN and CNN for HSI classification. Inspired by Mou \emph{et al.}\cite{mou2017deep}, we take all spectra of a pixel as sequence to train the RNN, which is good at exploiting spectral feature of HSI. Spatial feature is extracted with CNN after reducing dimension of HSI by principal component analysis (PCA). Based on the extracted spectral and spatial features, MDCPE chooses credible samples from each class to update labeled data. Then two networks are trained with new updated labeled data iteratively and co-decide the classification results. The three major contributions of the proposed method are listed as follows.

\begin{enumerate}[1)]
\item A semi-supervised co-training MDCPE is proposed in this paper, which presents two advantages: a) MDCPE enlarges insufficient labeled data by recovering unlabeled samples, which avoids of wasting unlabeled HSI information and enhances the robustness of extracted feature; b) k-means clustering is adopted to keep the updated data class-balanced, which helps the learners extract feature of all classes comprehensively and thus acquire a balanced accuracy across classes.

\item RNN and CNN cooperate as two learners of MDCPE to extract spectral-spatial feature of HSI. Taking all HSI spectra as sequential signals, RNN is expert in exploring relationship between spectra and extracting discriminative spectral feature. CNN is utilized to learn robust and effective spatial feature as its deep layer structure and superiority in learning of neighbor information.

\item In this paper, RNN and CNN are trained with MDCPE co-training from spectral and spatial views. Deep neural networks, intergraded with co-training method, not only adept in extracting robust and effective feature, but also promote accuracy and strengthen generation ability by making full use of unlabeled data.
\end{enumerate}

The remainder of this paper is organized as follows. Section II briefly introduces the preliminaries of related methods. The proposed method for HSI classification, including MDCPE co-training and two learners (RNN, CNN) are described in detail in section III. Network configurations, experimental results, and discussion are provided in section IV. Finally, section V concludes this paper briefly.

\section{Preliminaries}
In this section, we mainly recall background information of traditional DCPE co-training and RNN, CNN classification models.
\begin{figure}
	\centering
    \includegraphics[width=0.4\textwidth, height=0.3\textwidth]{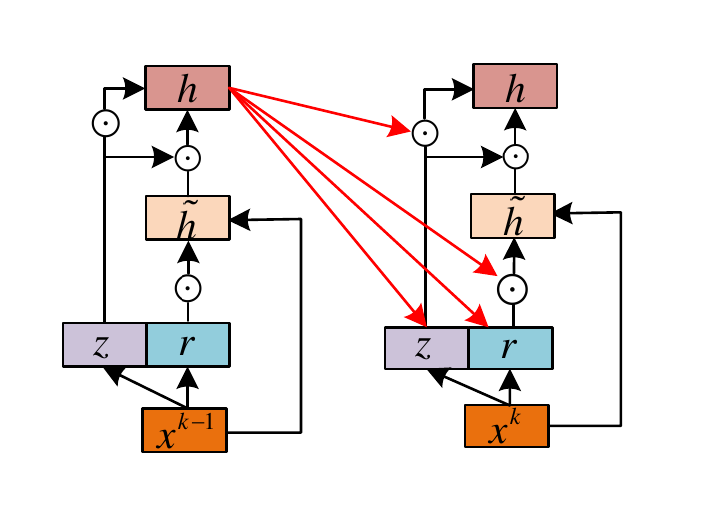}
    \caption{Graph model of GRU. \(z\) and \(r\) are the update gate and reset gate. The new memory content is denoted by \(\tilde{h}\).}
    \label{fig:GRU}
\end{figure}
\subsection{DCPE Co-training}
Co-training, as a complex generative branch of SSL model, is firstly proposed for classification of web pages\cite{blum1998combining}. In traditional co-training approach, two classifiers are trained separately based on two sufficient and redundant feature subsets (views), and then select unlabeled data with the most credible predicted labels as the new labeled data\cite{dopido2013semisupervised}.

DCPE is first proposed in\cite{xu2012dcpe}, in which the diversity based class probability estimation from two classifiers are utilized. The samples that have same labels and biggest difference class probability are chose to update the labeled data in DCPE. New recovered data has the decision agreement and the largest diversity in class probability estimation between two learning algorithms\cite{xu2012dcpe}.

Suppose that there are \(k\) classes \(c_i, i = 1,2, \cdots , k\) in classification issue. For each data \(x\), the class probability estimation for each class is defined as \(P(c_i|x)\). Data x will be allocated the most credible label, \(L(x) = argmax(P(c_i|x))\). In DCPE, the samples that have same predicted labels of two classifiers are allocated the predicted labels. Labels of the samples that have biggest difference class probability are calculated by
\begin{equation}
{label(x)} = \arg\max(P_2(c_i|x))-P_1(c_i|x))\label{8}
\end{equation}
\begin{equation}
{label(x)} = \arg\max(P_1(c_i|x))-P_2(c_i|x))\label{9}
\end{equation}
where \(P_1(c_i|x)\) and \(P_2(c_i|x)\) denote the class probability estimation for each category of classifier 1 and 2. Equations (\ref{8}), (\ref{9}) are utilized to update the training samples of classifier 1 and 2, respectively. When we ensemble two classifiers given each testing data during the testing process, the combined classifier computes the probability \(P(c_i|x)\) of class \(c_i\) via
\begin{equation}
{P(c_i|x)} = {P_1(c_i|x)P_2(c_i|x)}\label{10}
\end{equation}
Then the labels of the testing data is decided by \(arg max(P(c_i|x))\).

\subsection{Recurrent Neural Network}
RNN is a class of artificial neural network that extends conventional feedforward neural network with loops in connections\cite{mou2017deep}. RNN network is able to process sequential signal by using a recurrent hidden state that is activated at each step depending on the previous hidden state\cite{williams1989learning,rodriguez1999recurrent}. In this way, it can exploit temporal connection between input units and induce output of next step. RNN have gained significant attention for solving many challenging problems involving sequential data analysis, such as language modeling\cite{sundermeyer2015feedforward}, machine translation\cite{bahdanau2014neural}, and speech recognition\cite{graves2014towards,graves2013speech}.

In the traditional RNN model\cite{williams1989learning}, for a sequential input \(x=(x_1,x_2,\cdots,x_T)\), where \(x_i\) represents the input of \(ith\) time step. Its hidden vector sequence \(h=(h_1,h_2,\cdots,h_T)\) is calculated by:
\begin{equation}
{h_t} = \varphi(W_h \odot h_{t-1} + U_h \cdot h_{t-1})\label{1}
\end{equation}
where \(t\) ranges from 1 to \(T\). \(W_h\), \(U_h\) are coefficient matrixes for the input at present step and the state of hidden unit at previous step. \(\varphi (\cdot)\) is nonlinear activation function of hidden layer and \(y=(y_1,y_2,\cdots,y_T)\) is the output vector. In some cases such as classification of HSI, \(y_T\) is the only output of RNN as predicted label.

Gated recurrent unit (GRU) is introduced to learn long-term dependencies and alleviate gradient vanishing problem\cite{cho2014properties,gal2016theoretically}. The graph model of GRU is shown in Fig. \ref{fig:GRU}. Moreover, GRU has few parameters and is more suitable for small number of samples. By constructing some gates, GRU decides whether to update the cell state or to forget the former cell memory. Traditional RNN calculates hidden state by Equation (\ref{1}), which merely calculates the weighted sum of inputs and activates with a nonlinear function. However, a GRU-based recurrent layer calculates the activation of gated recurrent units by
\begin{equation}
{h_t} = z_t \odot h_{t-1} + (1-z_t) \odot {\tilde h_t}\label{2}
\end{equation}
where \(z_t\) denotes update gate that determines which part of the activation will be updated and \(\odot\) means an elementwise multiplication. \(h_{t-1}\) is the activation of gated recurrent units at previous step, and \(\tilde{h}_t\) represents candidate activation. The update gate is calculated by Equation (\ref{3}), and candidate activation is obtained by Equation (\ref{4}).
\begin{equation}
{z_t} = \sigma ({W_z}{x_t} + {U_z}{h_{t - 1}})\label{3}
\end{equation}
\begin{equation}
{\tilde h_t} = \tanh (W{x_t} + {r_t} \odot U{h_{t - 1}})\label{4}
\end{equation}
\begin{equation}
{r_t} = \sigma ({W_r}{x_t} + {U_r}{h_{t - 1}})\label{5}
\end{equation}
where \(\sigma(\cdot)\) denotes a logistic sigmoid function and \(\tanh(\cdot)\) is the hyperbolic tangent function. \(W\) and \(U\) terms are weight matrixes, and \(r_t\) represents the reset gate that determine whether to abandon the previous activation or not when generating candidate activation. \(r_t\) is calculated by Equation (\ref{5}).

\subsection{Convolutional Neural Network}
A complete CNN structure consists of convolutional layers and pooling layers, usually followed several fully connected layers and a softmax classification layer\cite{lecun1998gradient}. 3-D convolutional kernel is utilized in 3-D CNN\cite{chen2016deep}, which is often used to extract spectral and spatial features simultaneously in HSI classification models.

In the \(i\)th layer, the value of a neuron \(n_{ij}^{xyz}\) at the position of the \(j\)th feature map in 3-D CNN is calculated by Equation (\ref{6})
\begin{equation}
n_{ij}^{xyz} = f(\sum\limits_m {\sum\limits_h^{{H_i} - 1} {\sum\limits_l^{{L_i} - 1} {\sum\limits_d^{{D_i} - 1} {w_{ijm}^{hld}v_{(i - 1)m}^{(x + h)(y + l)(z + d)} + {b_{ij}}}}}})\label{6}
\end{equation}
where \(m\) denotes the feature maps in the (\(i-1\))th layer that are connected to the \(j\)th feature map. \(H_i\), \(L_i\) and \(D_i\) denote the height, the width and the dimension (along the spectral dimension) of the convolutional kernel. \(w_{ijm}^{hld}\) means the weight at the position (\(h,l,d\)) that is connected to the \(m\)th feature map, and \(b_{ij}\) is the bias of the \(j\)th feature map in the \(i\)th layer. \(f(\cdot)\) denotes nonlinear activation function, which is sigmoid activation function commonly:
\begin{equation}
f(x) = {(1 + {\exp ^{( - x)}})^{ - 1}}\label{7}
\end{equation}

\begin{figure}
	\centering
    \includegraphics[width=0.5\textwidth]{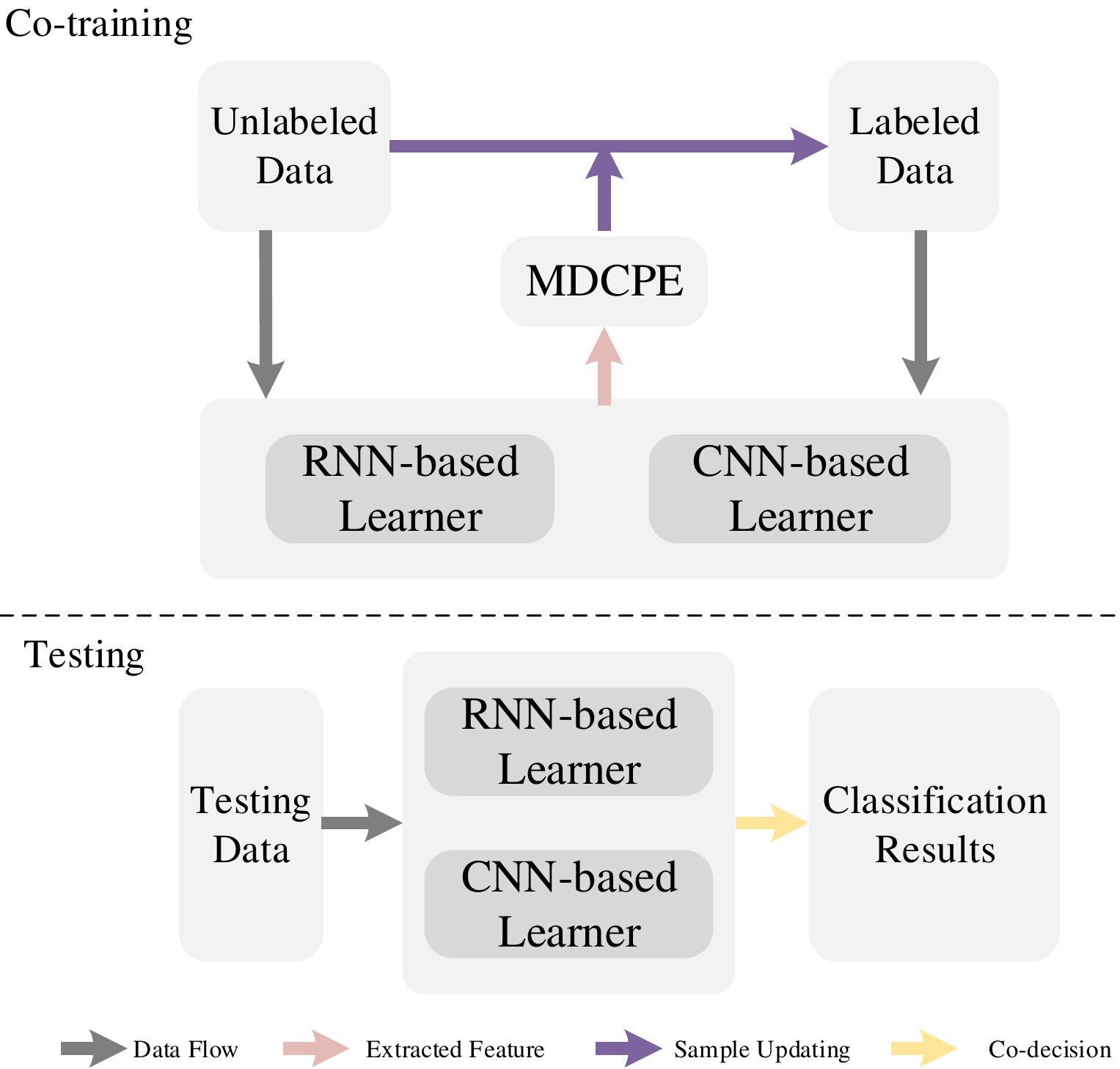}
    \caption{Flowchart of the proposed method.}
    \label{fig:flowchart}
\end{figure}

\section{Proposed Method} 
The flowchart of the proposed MDCPE co-training is illustrated in Fig. \ref{fig:flowchart}. During training process, as shown on top, RNN and CNN are pre-trained on labeled data and then are used to extract spectral-spatial feature of unlabeled data. Based on the feature, MDCPE is utilized to select the most credible unlabeled samples to replenish labeled data. New labeled data is used to train the classifiers again and loop this process until MDCPE meets the stop criteria. In addition, part of labeled data are utilized as validation data to monitor training process for saving the model with highest validation accuracy. During testing process, as shown in bottom of Fig. \ref{fig:flowchart}, classification results are acquired by co-decision of co-trained RNN and CNN with Equation (\ref{10}).

It is obvious that three sections play important roles in the proposed method: MDCPE co-training, RNN-based spectral feature learner and CNN-based spatial feature learner. We will introduce our innovative work on three parts in section III-A, III-B and III-C, respectively.

\subsection{MDCPE Co-training}
Traditional DCPE is a single view method, which is unnecessary to split feature to different groups. Rather than learning on the same feature, MDCPE expands DCPE to spectral and spatial views to extract discriminative spectral-spatial feature with RNN and CNN, which enhances effectiveness of extracted feature and promotes HSI classification results.

Furthermore, traditional DCPE updates labeled data without considering unbalance between classes in updated data, which diminish the comprehensiveness of extracted features and the overall accuracy. The classes that are allocated too many training samples will be over-trained and get high accuracies, while the performance is poor when the classes have inadequate updated samples. In MDCPE, k-means clustering is combined with DCPE to ensure the balance in updated data. The most credible samples selected to replenish labeled data are determined by results of two learners and k-means clustering.

MDCPE co-training is described in the Algorithm \ref{alg:MDCPEalg} in detail. Different learners have diverse calculation mechanisms of class probability. In MDCPE co-training, we utilize Softmax layers of deep neural network (RNN, CNN) to estimate the class probability.

\makeatletter
\newenvironment{breakablealgorithm}
  {\begin{center}
     \refstepcounter{algorithm}
     \hrule height.8pt depth0pt \kern2pt
     \renewcommand{\caption}[2][\relax]{
       {\raggedright\textbf{\ALG@name~\thealgorithm} ##2\par}%
       \ifx\relax##1\relax 
         \addcontentsline{loa}{algorithm}{\protect\numberline{\thealgorithm}##2}%
       \else 
         \addcontentsline{loa}{algorithm}{\protect\numberline{\thealgorithm}##1}%
       \fi
       \kern2pt\hrule\kern2pt
     }
  }{\kern2pt\hrule\relax
   \end{center}
  }
\makeatother
\begin{breakablealgorithm}
    \caption{The proposed MDCPE co-training.}
    \label{alg:MDCPEalg}
    \begin{algorithmic}[1]
        \Require
            \(C_1\): learner 1; \(C_2\): learner 2; \(D_t\): labeled data set; \(D_u\): unlabeled data set; \(c_i\): class \(c_{i,i = 1,2,\cdots , k}\)
        \Ensure
            \(\widetilde{C_1}\): co-trained learner 1; \(\widetilde{C_2}\): co-trained learner 2
        \State allocate \(S_1 = D_t\), \(S_2 = D_t\);
        \State train \(C_1\), \(C_2\) on \(S_1\), \(S_2\);
        \State record predicted label, class probability estimation and extracted feature of two learners as \(L_1(\cdot)\), \(L_2(\cdot)\) , \(P_1(\cdot)\), \(P_2(\cdot)\) and \(F_1(\cdot)\), \(F_2(\cdot)\);
        \State record predicted label of k-means clustering as \(L_{km}(\cdot)\);
        \Repeat
            \State choose \(u_s\),\(u_{1m}\),\(u_{2m}\) from \(D_u\) to satisfy equations (\ref{11}), (\ref{12}) and (\ref{13});
            \State select one sample belonged to class \(c_i\) from \(D_t\) randomly to form cluster center \(O=\{o_i\}\), \(i={1,2,\cdots,k}\);
            \State centered on \(F_1(O)\), cluster \(F_1(u_s)\), \(F_1((u_{1m})\) with k-means clustering;
            \State similarly, cluster \(F_2(u_s)\), \(F_2((u_{2m})\), centered on \(F_2(O)\);
            \State update \(S_1 = S_1 + D_1\), which satisfies equations (\ref{14});
            \State update \(S_2 = S_2 + D_2\), which satisfies equations (\ref{17});
            \State eliminate \(D_1\), \(D_2\) from \(D_u\);
            \State train \(C_1\), \(C_2\) on new labeled data \(S_1\), \(S_2\).
        \Until \(D_u=\varnothing\)
    \end{algorithmic}
\end{breakablealgorithm}

In this algorithm, we predict all unlabeled data in each iteration rather than choosing some samples pool as DCPE does, because we want to enlarge the searching range and ensure there are enough samples in each class to be selected. During selecting samples to update labeled data, the sample \(u_s\) that satisfies Equation (\ref{11}) is selected.
\begin{equation}
{L_1(u_s)= L_2(u_s)}\label{11}
\end{equation}
Then we need to select samples that have highest diversity of class probability. When updating samples of RNN, labels of samples that have diversity of class probability estimation(CPE) are calculated by
\begin{equation}
{L_1(u_{1m})} = \arg\max(P_2(u_{1m})-P_1(u_{1m}))\label{12}
\end{equation}
Similarly, when updating samples of CNN, labels of the samples are calculated by
\begin{equation}
{L_2(u_{2m})} = \arg\max(P_1(u_{2m})-P_2(u_{2m}))\label{13}
\end{equation}
In order to keep classes of updated samples balanced, we utilize k-means clustering to classify \(F_1(u_s)\), \(F_1((u_{1m})\), \(F_2((u_{2m})\), from which we select fixed number of data from each class to update the labeled data. For RNN, updated data set \({D_1}\) that owns same labels and highest CPE is calculated by
\begin{equation}
{D_1} = \{d_{1s_i}\}\bigcup\{d_{1m_i}\}\label{14}
\end{equation}
where \(i=1,2,\cdots,k\). \(d_{1s_i}\) and \(d_{1m_i}\) are selected to update the labeled data of RNN learner and their labels are decided by
\begin{equation}
{L_1(d_{1s_i})} = L_{km}(d_{1s_i}) = c_i\label{15}
\end{equation}
\begin{equation}
{L_1(d_{1m_i})} = L_{km}(d_{1m_i}) = c_i\label{16}
\end{equation}
When selecting updated data for CNN, samples in \({D_2}\) data set are supposed to satisfy the follows:
\begin{equation}
{D_2} = \{d_{2s_i}\}\bigcup\{d_{2m_i}\}\label{17}
\end{equation}
where \(i=1,2,\cdots,k\) and \(d_{2s_i}\), \(d_{2m_i}\) are computed as follows:
\begin{equation}
{L_2(d_{2s_i})} = L_{km}(d_{2s_i}) = c_i\label{18}
\end{equation}
\begin{equation}
{L_2(d_{2m_i})} = L_{km}(d_{2m_i}) = c_i\label{19}
\end{equation}

Rather than choosing cluster centers randomly, we select one labeled sample from each class, denoted as \(O\) and extract spectral feature \(F_1(O)\) and spatial feature \(F_2(O)\) as clustering centers of two classifiers. By adding prior labeled information of classes, we alleviate the influence of choosing initial cluster centers randomly and thus can get better clustering performance. After clustering, choosing the samples whose predicted labels of classifiers are same with results of k-means clustering from each class to update labeled data. It's important to note that the selected unlabeled data will be remove from original unlabeled data set. New recovered data selected by the MDCPE has the decision agreement and the largest diversity in class probability estimation between two classifiers. In addition, updated data is balanced between classes, which contributes to improve overall classification results.

\subsection{RNN-based Spectral Feature Learner}
\begin{figure}[htbp]
	\centering
    \includegraphics[width=0.5\textwidth]{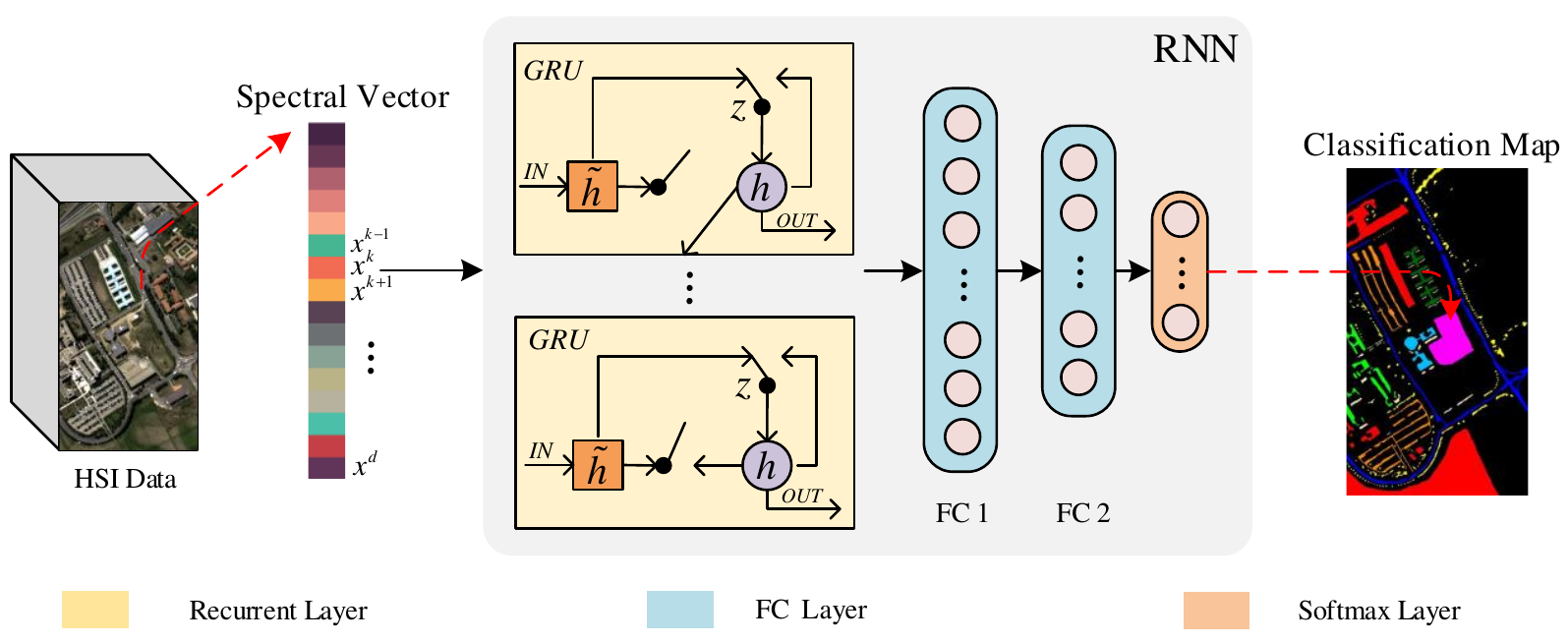}
    \caption{RNN-based spectral feature learner.}
    \label{fig:RNNflow}
\end{figure}

Since the temporal variability of a sequential signal is similar to the spectral variability of a hyperspectral pixel, the same idea can be transferred to hyperspectral pixel vector\cite{mou2017deep}.  Considering all spectra of a hyperspectral pixel as a sequence, we design a RNN learner, as shown in Fig. \ref{fig:RNNflow}, to extract spectral feature of HSI. Input of the classifier is a spectral vector of a HSI pixel \(x\), where its \(k\)th spectral band is defined as \(x^k\). Output is the predicted label of pixel \(x\). The detailed process is summarized as follows.

\begin{enumerate}[1)]
\item Feed \(x^k\) into the input layer of network.
\item The recurrent layer calculates the hidden state of current spectral band \(k\) after receiving the \(x^k\) and saves this information.
\item Next band \(x^{k+1}\) is input to recurrent layer. In order to calculate the activation of next \(k+1\) spectral band, the candidate activation and the activation of previous band \(k\) was linear summed with a controlling parameter of the update gate.
\item Train the learner on labeled training data.
\item Extract spectral feature of training data from FC2 layer after removing the Softmax layer.
\item Input testing pixel sequences into the learner to get classification results.
\end{enumerate}

Considering of insufficient labeled training data and huge data complexity in hundreds of spectral bands, the proposed RNN with GRU classifier has fewer parameters than traditional LSTM unit so it can alleviate overfitting. In addition, by using gates, the GRU will preserve the errors that can be back propagated through sequences and layers. Hence, RNN can learn over many bands of hyperspectral pixels without the risk of the vanishing gradient\cite{mou2017deep}. Moreover, as a learner of the proposed MDCPE co-training, RNN is good at capturing intrinsic characteristic and extract discriminative spectral feature of HSI.
\subsection{CNN-based Spatial Feature Learner}
\begin{figure}[htbp]
	\centering
    \includegraphics[width=0.5\textwidth]{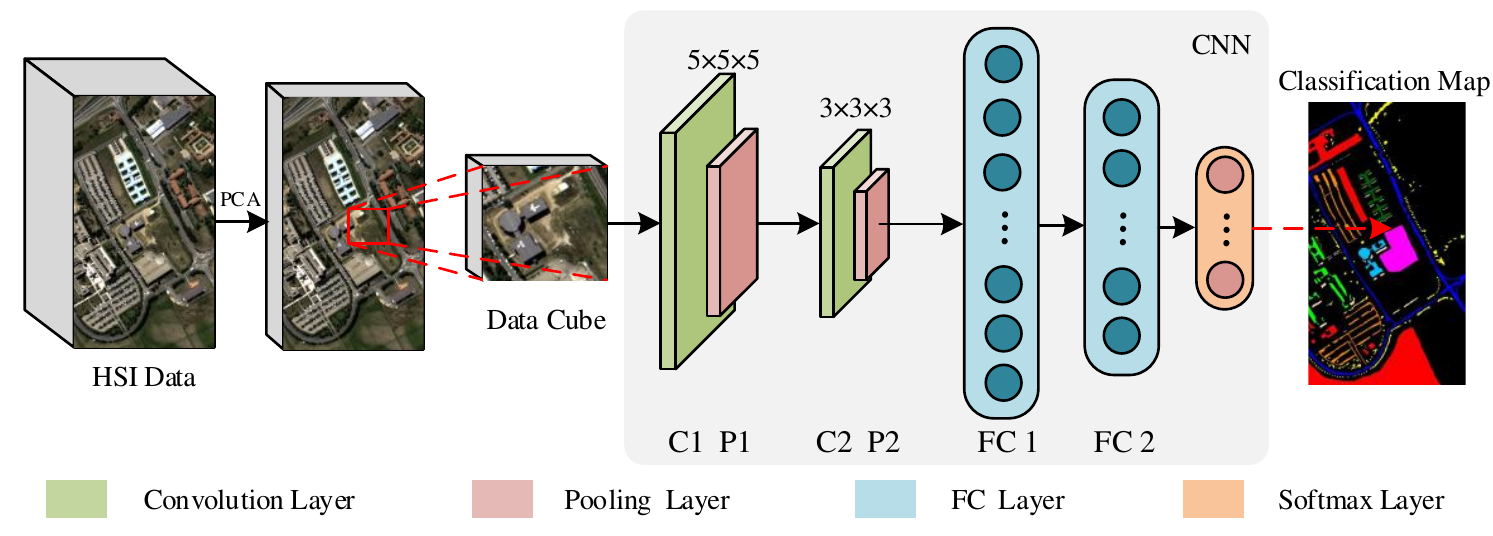}
    \caption{CNN-based spatial feature learner.}
    \label{fig:CNNflow}
\end{figure}

CNN is adopted as another learner of MDCPE co-training to learn effective spatial feature of HSI. Architecture of this learner is shown in Fig. \ref{fig:CNNflow}. The size of 3-D convolutional kernels of C1 and C2 layers are set as \(5\times5\times5\), \(3\times3\times3\), respectively. Strides of max pooling layers P1, P2 are 2. The fully connected layer FC1 owns 1024 units and the number of units FC2 has is equal to the classes of HSI data. Inputting data of CNN is generated by segmenting data blocks with fixed size after reducing dimensions of HSI by PCA. Spatial feature is extracted form FC2 layer by dismissing the Softmax layer of CNN learner.

As our CNN learner mainly aims at extracting robust spatial feature, dimensions of HSI is first reduced by principal component analysis (PCA), which could decrease the spectral redundancy and save computational time with a large margin\cite{zhao2016spectral}. Furthermore, Credibility of samples selected to update labeled data has a big impact on co-training of learner and thus classification accuracy. The 3-D CNN, as a base learner of MDCPE, can extract more discriminative spatial feature and enhance the confidence of updated samples.
\section{Experiment Results and Discussion}
In this section, we introduce three data sets used in our experiment and the configuration of the proposed MDCPE co-training. In addition, classification results based on the proposed method and other comparative methods are presented. All the experiments are conducted on HP z840 workstation equipped with an Intel Xeon E5 with 2.40 GHz and NVIDIA Titan X GPU. TensorFlow 1.2.0 and Keras 2.1.3 are utilized as software platforms.

\subsection{Data Description}
\subsubsection{Salinas Data (SA)} SA was collected by the 224-band AVIRIS sensor and is characterized by 3.7-meter spatial resolution. 20 water absorption bands were discarded, with 204 bands left to be applied in the experiment. This data set includes 16 classes ground-truths, with the size of \(512\times217\) pixels. Table \ref{tab:data1} lists all classes and their corresponding training and test samples.
\begin{table}[htbp]
\centering
\caption{NUMBER OF TRAINING AND TESTING SAMPLES IN SALINAS DATA SET}
\label{tab:data1}
\includegraphics[width=0.5\textwidth]{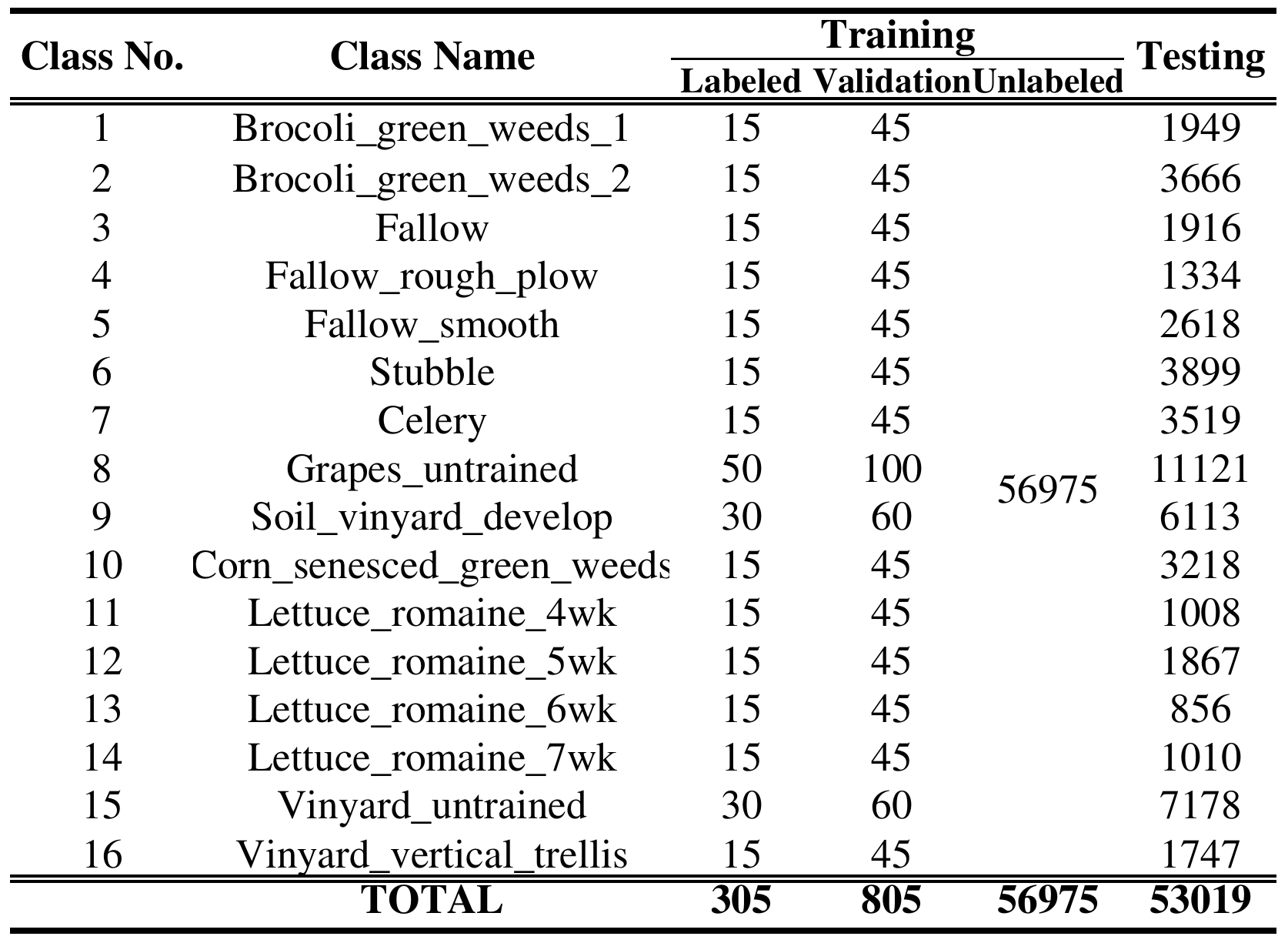}
\end{table}

\subsubsection{Pavia University (PU)} This data set is acquired by reflective optics system imaging spectrometer (ROSIS). The ROSIS-03 sensor comprises 115 spectral channels ranging from 430 to 860\(nm\). Once 12 noisy channels have been removed, the remaining 103 spectral channels are adopted in this paper. The image has \(610\times340\) pixels with a spatial resolution of 1.3\(m\) per pixel. The available training samples of this data set contains nine urban land-cover types. Table \ref{tab:data2} provides information about different classes and corresponding training and test samples.

\subsubsection{Pavia Center (PC)} PC is also acquired by ROSIS. In this data set, 13 noisy channels have been removed and the remaining 102 spectral channels are utilized in this paper. The image is of \(1096\times715\) pixels covering the center of Pavia, which spatial resolution is 1.3 \(m\) per pixel. The available training samples contains nine urban land-cover classes and Table \ref{tab:data3} provides information about training and test samples of different classes.
\begin{table}[htbp]
\centering
\caption{NUMBER OF TRAINING AND TESTING SAMPLES IN PAVIA UNIVERSITY DATA SET}
\label{tab:data2}
\includegraphics[width=0.5\textwidth]{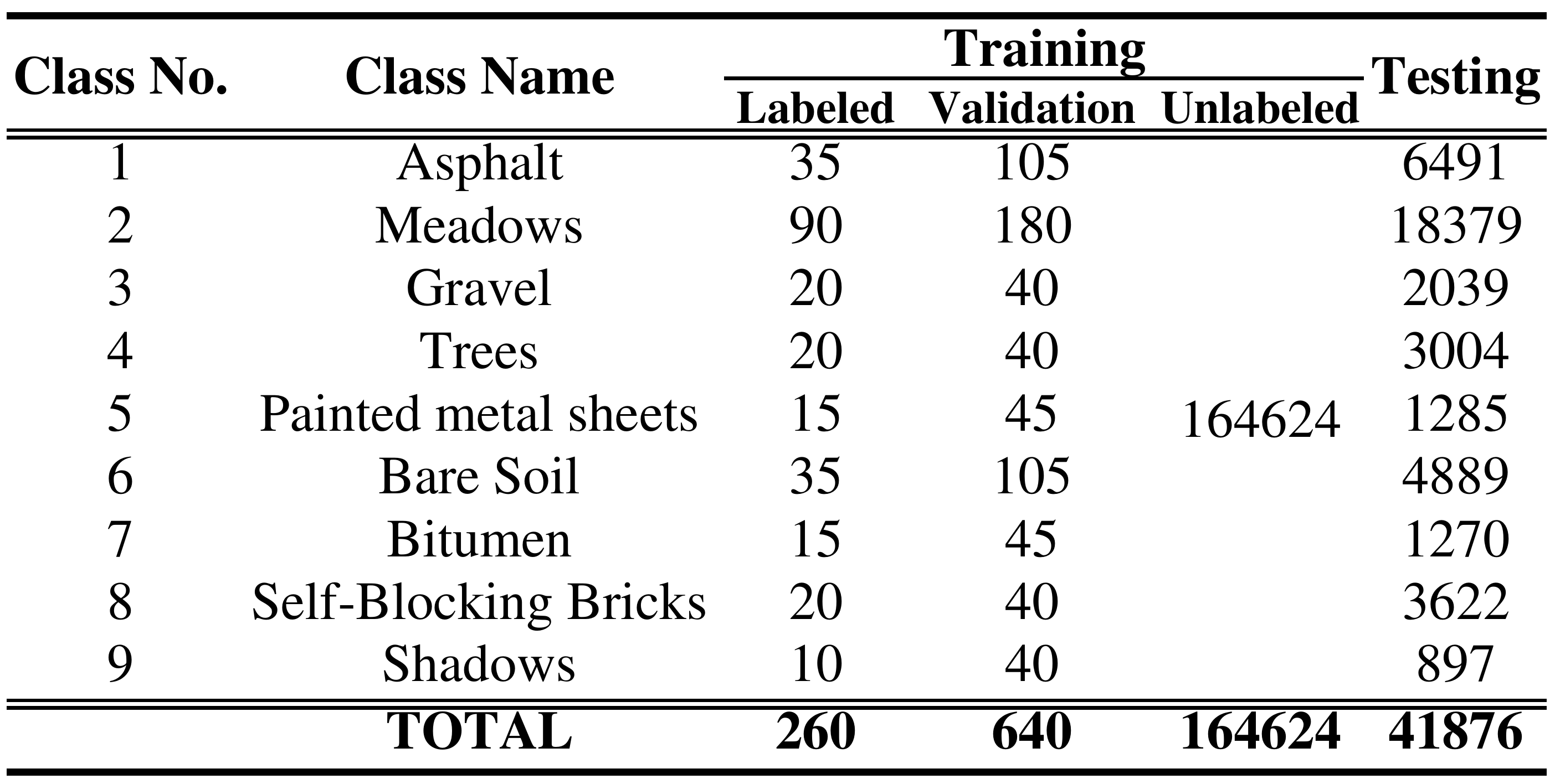}
\end{table}

We mainly utilize the three data sets mentioned above to evaluate the performance of the proposed method. These data sets are selected for several reasons: firstly, they both have high spatial resolution and cover wide range of real cases, which make the method difficult to classify; secondly, they have different scenery: SA data set is obtained in a mixed vegetation site, while PU and PC data sets are acquired in an urban site. These characters facilitate us to assess the effectiveness and robustness of the proposed method.
\begin{table}[htbp]
\centering
\caption{NUMBER OF TRAINING AND TESTING SAMPLES IN PAVIA CENTER DATA SET}
\label{tab:data3}
\includegraphics[width=0.5\textwidth]{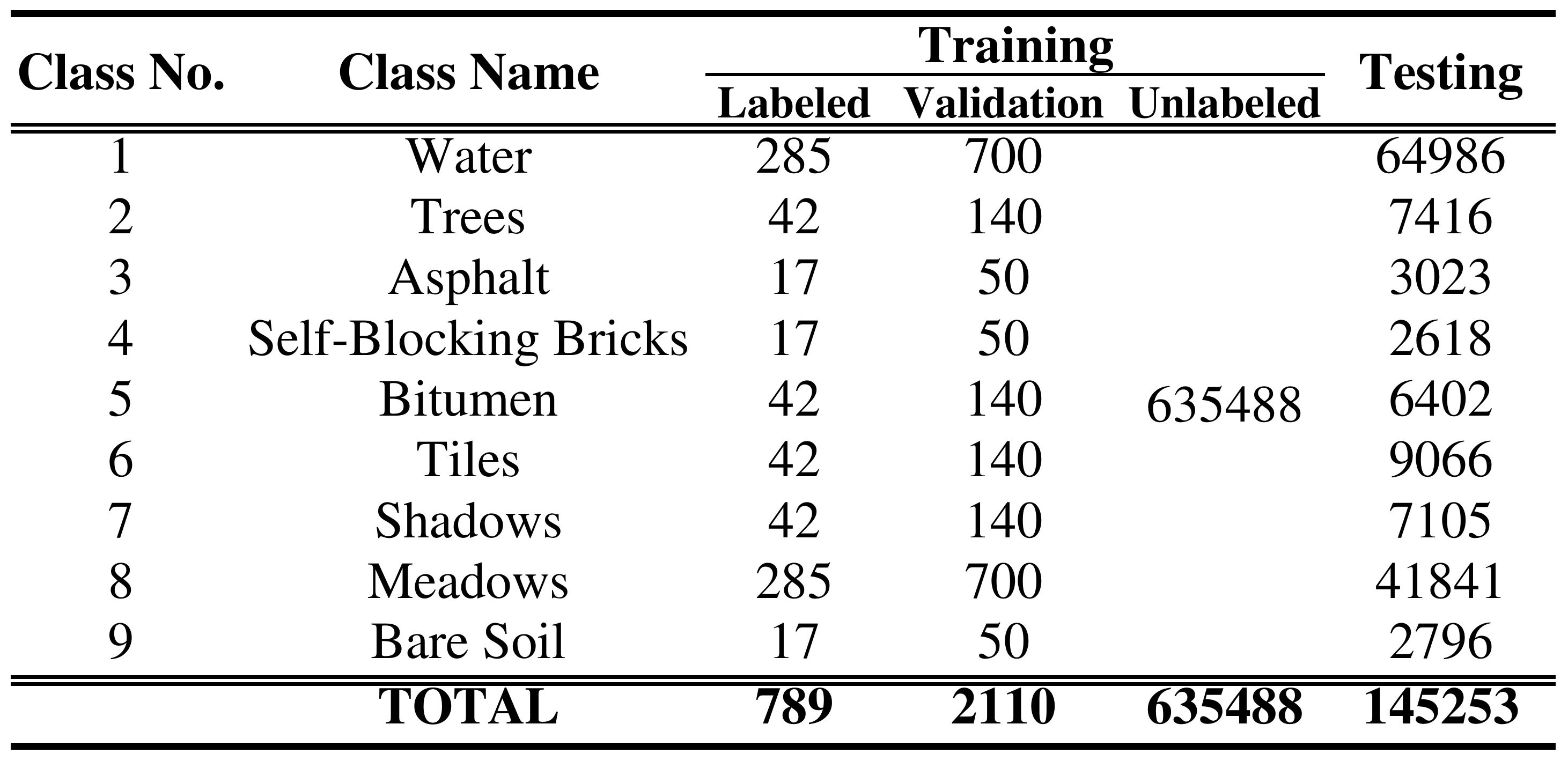}
\end{table}

\begin{figure}[htbp]
	\centering
    \includegraphics[width=0.45\textwidth]{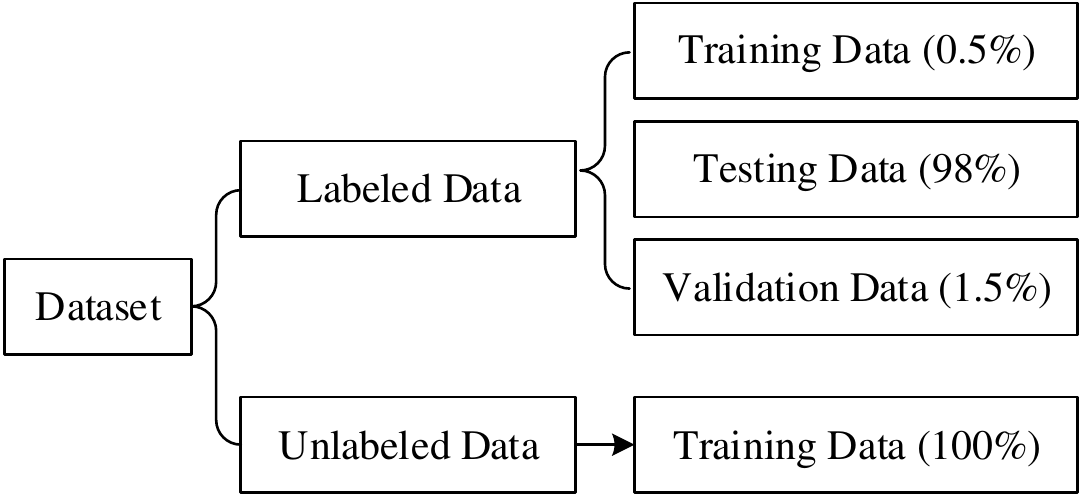}
    \caption{Data splitting configuration for experiment.}
    \label{fig:data}
\end{figure}

In our experiment, data set is split randomly into two groups as Fig. \ref{fig:data} shown, training data and testing data (98 \% of given labeled data). Training data consists of three parts, labeled samples (about 0.5\% of given labeled data), all unlabeled samples and validation data (1.5\% of given labeled data).

In SA data set, we choose 3 principal components that could consist of 99\% information of the original HSI, and then extract image blocks of when extracting spatial feature based on CNN. Similarly, 4 principal components are chosen in PU data set and 3 components for PC data set. In addition, all training and testing samples are standardized by min-max normalization. In the \(k\)th spectral channel, the value of a pixel after normalization \({x^{'}}_{ij}^k\) at the position of the \(i\)th line, \(j\)th column is calculated by:
\begin{equation}
{x^{'}}_{ij}^k = \frac{x_{ij}^{k}-\min(x_{ij}^{k})}{{\max(x_{ij}^{k})-\min(x_{ij}^{k})}}\label{19}
\end{equation}
where \(\min(\cdot)\) and \(\max(\cdot)\) means the minimum and maximum in all pixels of \(k\)th spectral channel, respectively.

\subsection{Parameter Setting}
As two neural network learners of MDCPE co-training, RNN and CNN are all trained by stochastic gradient descent (SGD) algorithm. Because labeled data sets are small, 32 is set as the batch size of two classifiers for SA and PU data sets, and 64 is chose for PC data set. Then we analyze several parameters that impact training process and classification accuracy, such as learning rate, patch size of input block, and iterations of MDCPE, etc. The models with the highest classification accuracy on validation data were saved during the training process of each configuration.

\subsubsection{Learning rate} Learning rate controls the learning step during training process. Too big learning rate will cause vibration and tend to converge on local optimum, while network converges too slow if it too small. The learning rate of RNN and CNN are chose from \([0.01, 0.008, 0.005, 0.003, 0.001, 0.0006, 0.0003]\) and \([0.001, 0.0008, 0.0005, 0.0003, 0.0001, 0.00006, 0.00003]\), respectively. Based on the classification accuracy, the optimal learning rate of RNN and CNN for SA data set are 0.001 and 0.0003, 0.003 and 0.0001 for PU data set and 0.001 and 0.0001 for PC data set. During the co-training of two learners with MDCPE, the learning rate is set as optimal learning rate and remains unchanged.

\begin{table}[htbp]
\centering
\caption{OA OF CNN WITH DIFFERENT INPUT SIZES}
\label{tab:spatialsize}
\includegraphics[width=0.45\textwidth]{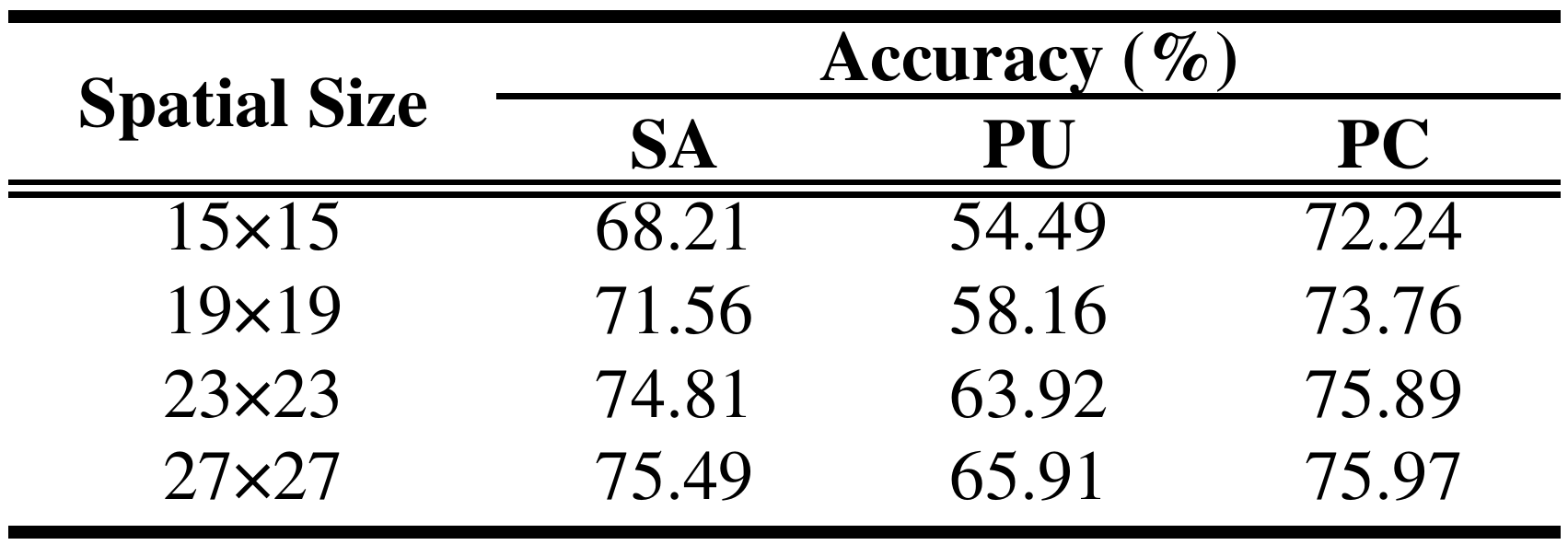}
\end{table}

\subsubsection{Patch size} CNN, as a learner of MDCPE, is utilized to extract spatial feature and its performance badly depends on the size of input block. In order to evaluate the influence of spatial size of input blocks during training of CNN, we set the size of input block \(15\times15\), \(19\times19\), \(23\times23\), \(27\times27\) to extract discriminative spatial feature. Table \ref{tab:spatialsize} shows classification result on two validation data sets. These results are acquired in 500-epoch training processes for each setting in three data sets. In all data sets, the classification results increase with the spatial size of input cubes, because larger input blocks are, more spatial information is offered. Hence, we fixed the spatial size of input cube to make a fair comparison between different classification methods.
\begin{table}[htbp]
\centering
\caption{OA OF CNN WITH DIFFERENT REGULARIZERS. THE BEST ACCURACY IN EACH ROW IS SHOWN IN BOLD}
\label{tab:dropout}
\includegraphics[width=0.45\textwidth]{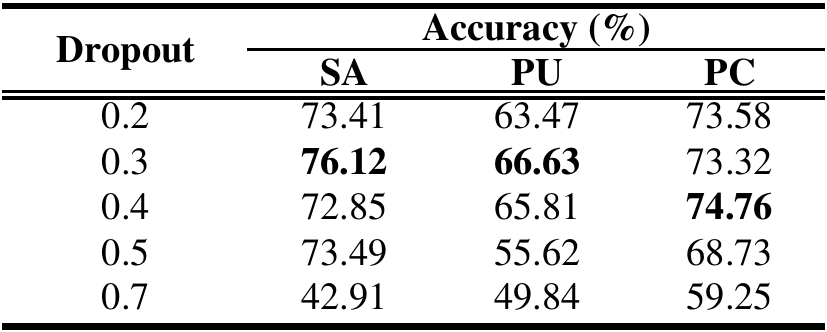}
\end{table}

\subsubsection{Dropout} Dropout is adopted to regularize the CNN training process and keep from overfitting. Table \ref{tab:dropout} shows several groups of evaluation results with different dropout proportion. These results are also acquired in 500-epoch training processes for each setting in three data sets. As Table \ref{tab:dropout} shows, 30\% dropout is selected for SA and PU data sets and 40\% for PC data set to get the highest accuracy.

\begin{figure}[htbp]
	\centering
    \includegraphics[width=0.5\textwidth]{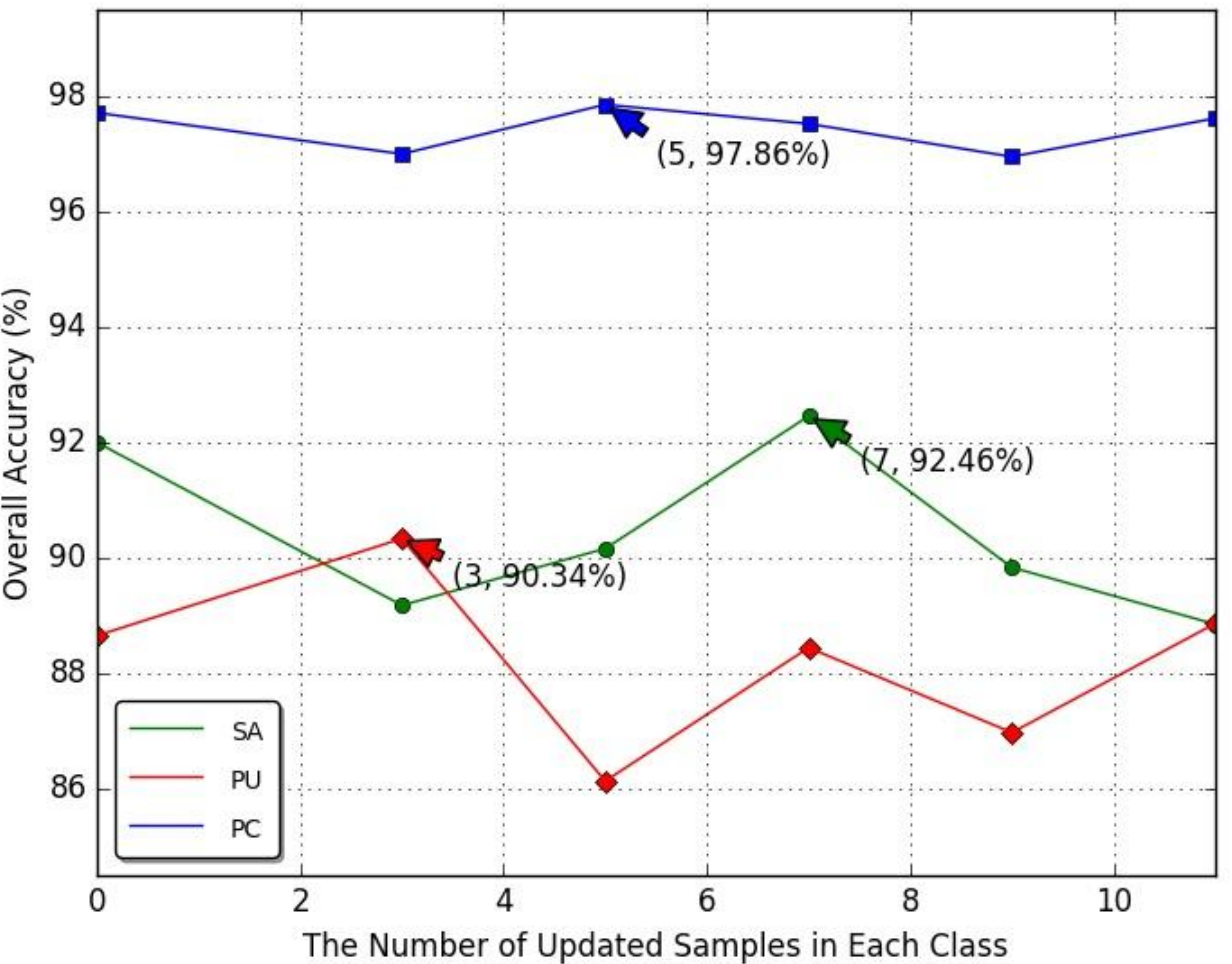}
    \caption{OA (\%) of MDCPE with different numbers of updated sample.}
    \label{fig:update}
\end{figure}

\subsubsection{Number of updated samples in each class} In the analysis of co-training\cite{wang2007analyzing}, it claims that the size of recovering data in each iteration should not affect the original data distribution. So in our experiment, the number of new labeled sample in each iteration is fixed and far less than the original number of labeled samples. MDCPE is adopted to select the unlabeled sample from each class and recover these samples with a predicted label. These updated data, with the initial labeled data together, are used to train the learners in next iteration. Fig. \ref{fig:update} presents performance of different number of updated samples in each class. These results are acquired in 5 iterations with MDCPE for each setting in three valid data sets. As Fig. \ref{fig:update} shows, In SA data set, 7 samples are selected from each class in every iteration of MDCPE co-training, 3 samples in PU data set and 5 samples in PC data set.

\begin{figure}[htbp]
	\centering
    \includegraphics[width=0.5\textwidth]{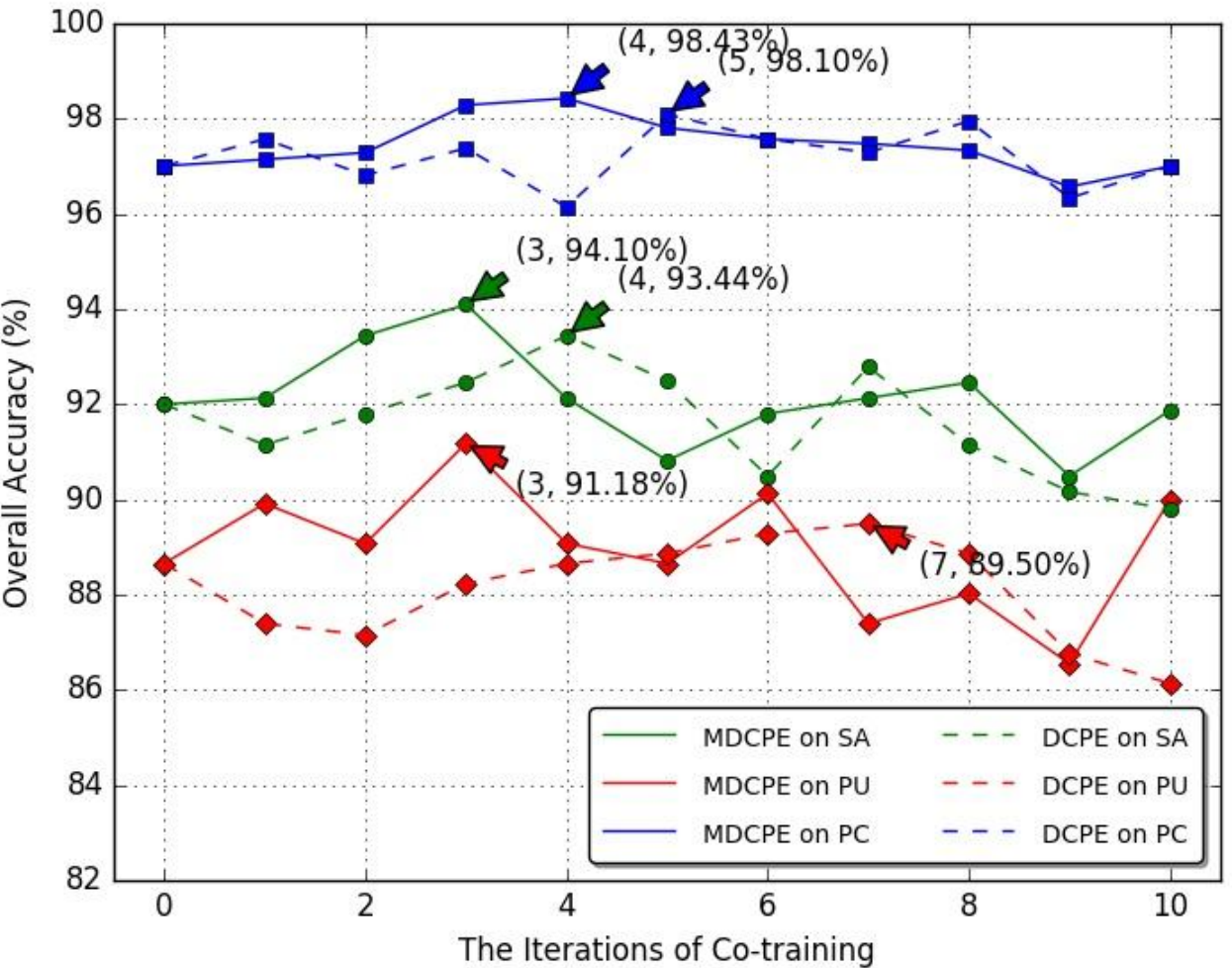}
    \caption{OA (\%) of MDCPE and DCPE with different iterations.}
    \label{fig:iter}
\end{figure}

\begin{table*}[htbp]
\centering
\caption{CLASSIFICATION RESULTS OF DIFFERENT METHODS FOR THE SA DATA SET. THE BEST ACCURACY IN EACH ROW IS SHOWN IN BOLD}
\label{tab:SA_acc}
\includegraphics[width=0.95\textwidth]{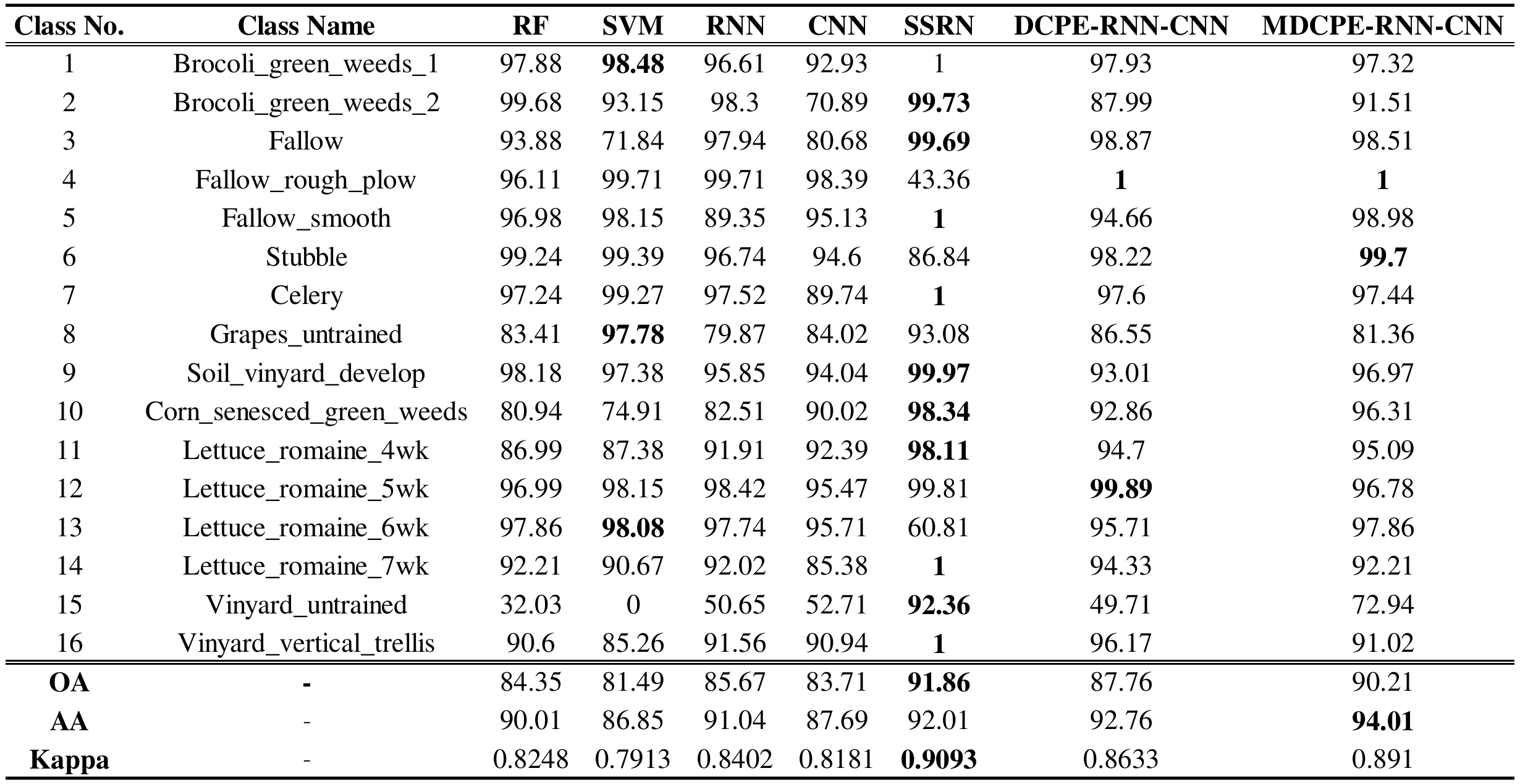}
\end{table*}

\subsection{Classification Results}

\begin{table*}[htbp]
\centering
\caption{CLASSIFICATION RESULTS OF DIFFERENT METHODS FOR THE PU DATA SET. THE BEST ACCURACY IN EACH ROW IS SHOWN IN BOLD}
\label{tab:PU_acc}
\includegraphics[width=0.95\textwidth]{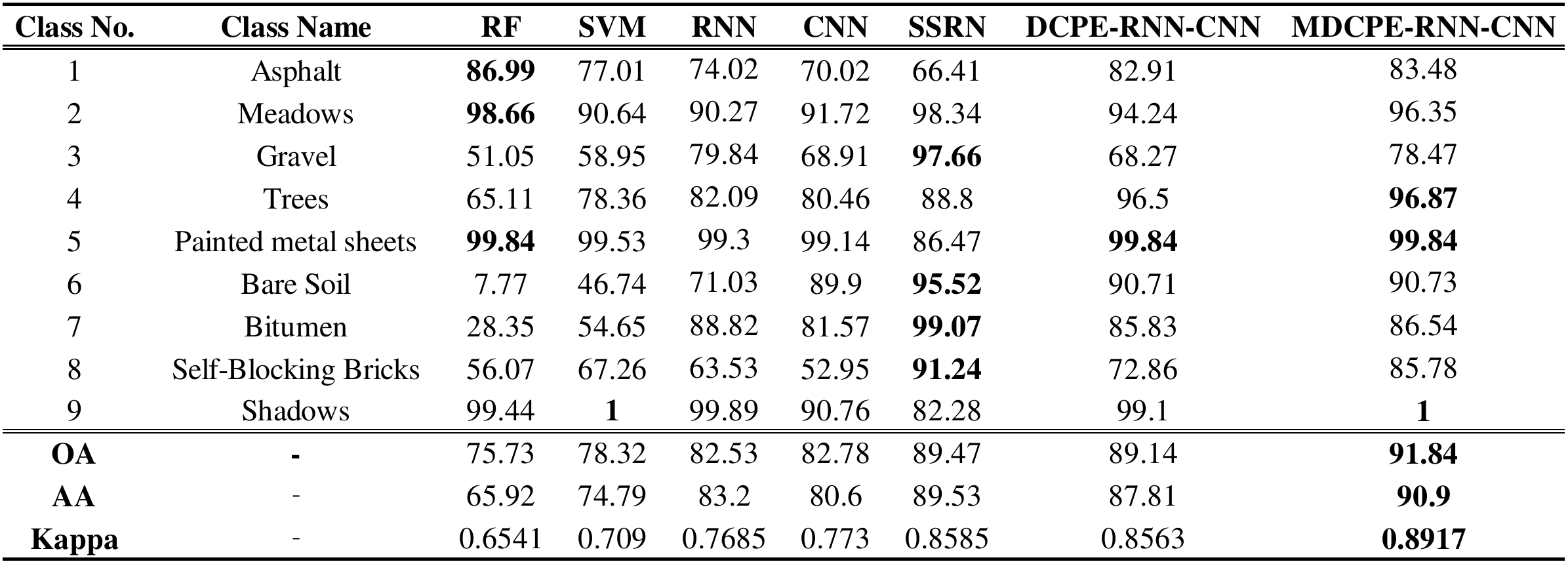}
\end{table*}
\subsubsection{Iterations} In the proposed MDCPE co-training, as the fixed number of samples are selected from unlabeled data to update the labeled data, the unlabeled data is decreasing with the increase of iterations. If there are not enough unlabeled data to replenish the labeled data, the semi-supervised learning tend to be overfitting\cite{schuurmans2002metric}. So we set the maximum of iterations to control the training process and prevent overfitting. Co-training process will not stop until the iterations equal to the maximum of iteration. However, last iteration is not always the best result\cite{xu2012dcpe}. So, validation data is used to control the iterations and monitor the training process in order to end in highest accuracy. In each iteration, we calculate the accuracy of validation data based on trained learner and save the learner with highest accuracy for next training iteration. Accuracy of validation data in each iteration is shown in Fig. \ref{fig:iter}, in which the performance of DCPE and MDCPE are shown together. The final output learners, which have the highest accuracy among all iterations on validation data, are selected and applied on testing data. As shown inn Fig. \ref{fig:iter}, The highest accuracy of DCPE and MDCPE for SA data set occurs in 4th and 3rd iterations, 7th and 3rd iterations for PU data set, 5th and 4th iterations for PC data set, respectively.

\subsubsection{Comparative Methods} To evaluate effectiveness of proposed method, we compared it with vector-based models: random forest (RF), support vector machine (SVM) and state-of-the-art deep learning models, such as RNN, CNN and SSRN. Moreover, in order to validate the superiority of MDCPE, we design DCPE-RNN-CNN, a semi-supervised classification model based on DCPE co-training of CNN and RNN learners. The contrastive methods are summarized as follows.
\begin{figure*}[htbp]
	\centering
	\subfigure[]{\includegraphics[width=0.18\textwidth]{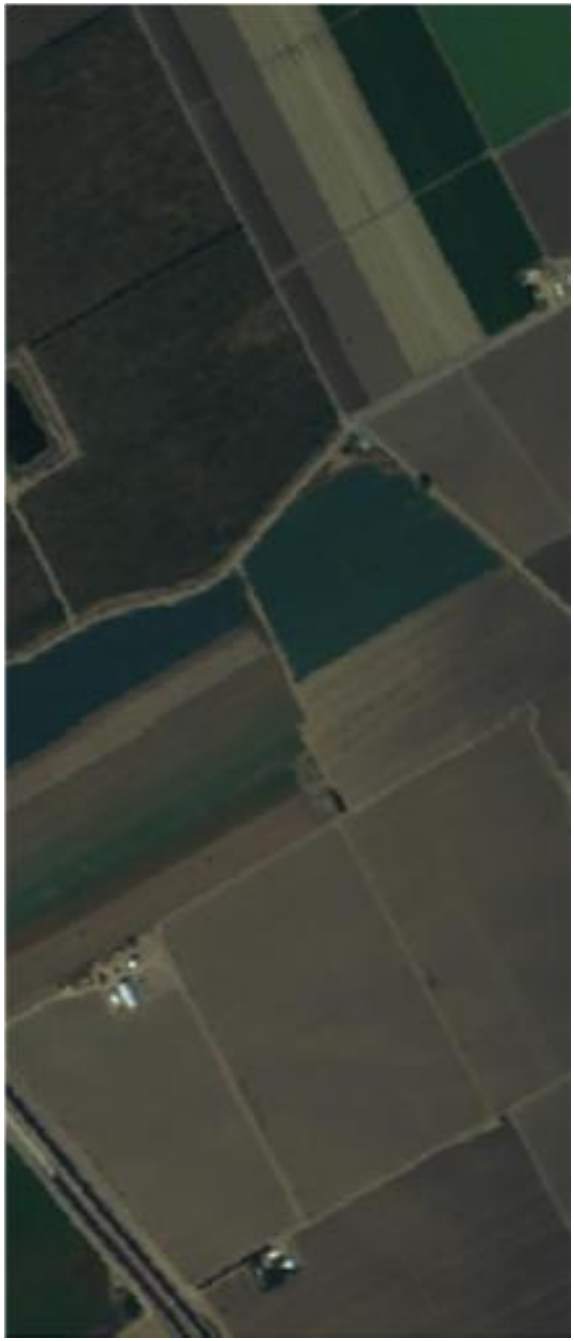}\label{fig:FCI}}
	\hspace{0.2cm}
	\subfigure[]{\includegraphics[width=0.18\textwidth]{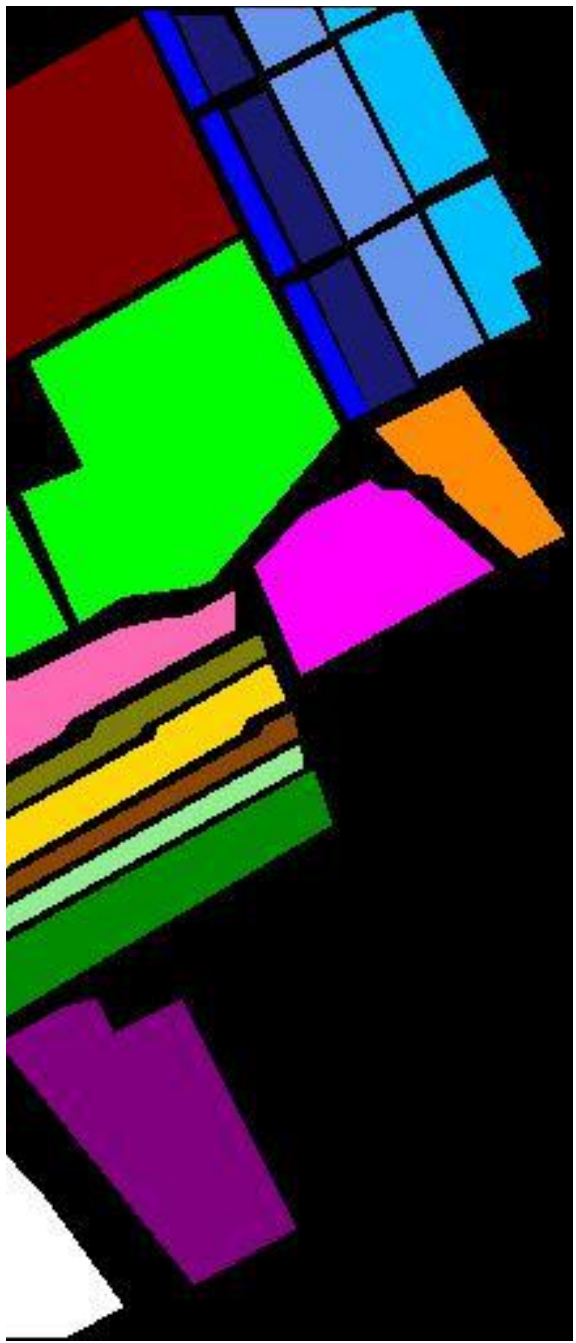}\label{fig:GT}}
	\hspace{0.2cm}
	\subfigure[]{\includegraphics[width=0.18\textwidth]{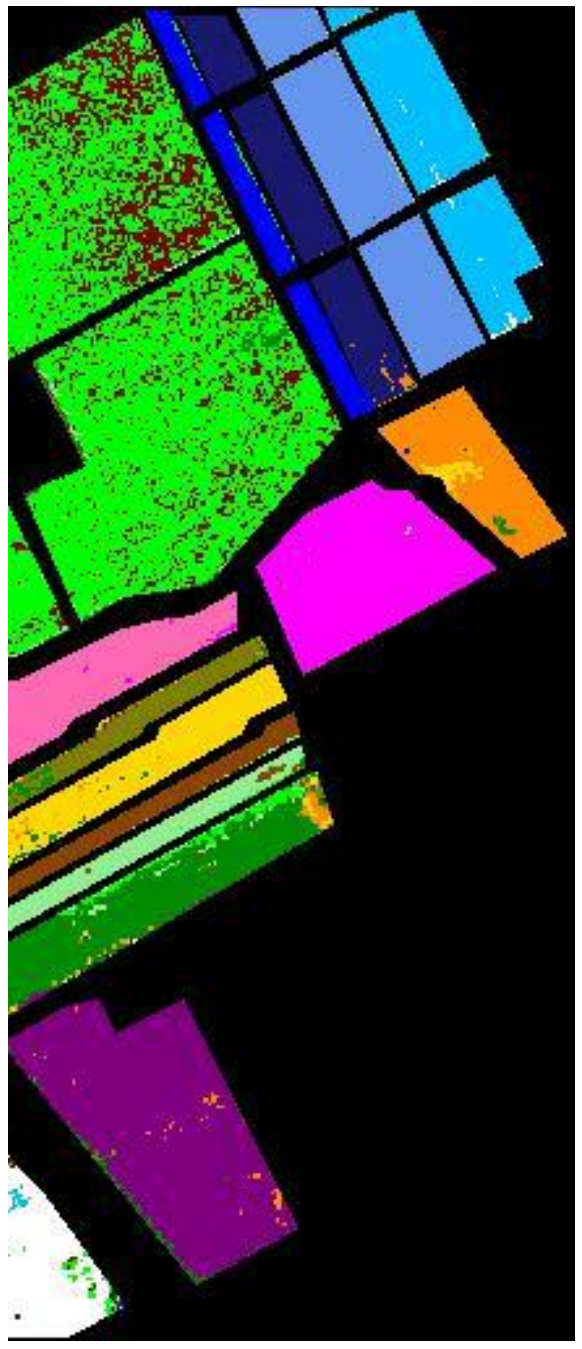}\label{fig:RF}}
	\hspace{0.2cm}
	\subfigure[]{\includegraphics[width=0.18\textwidth]{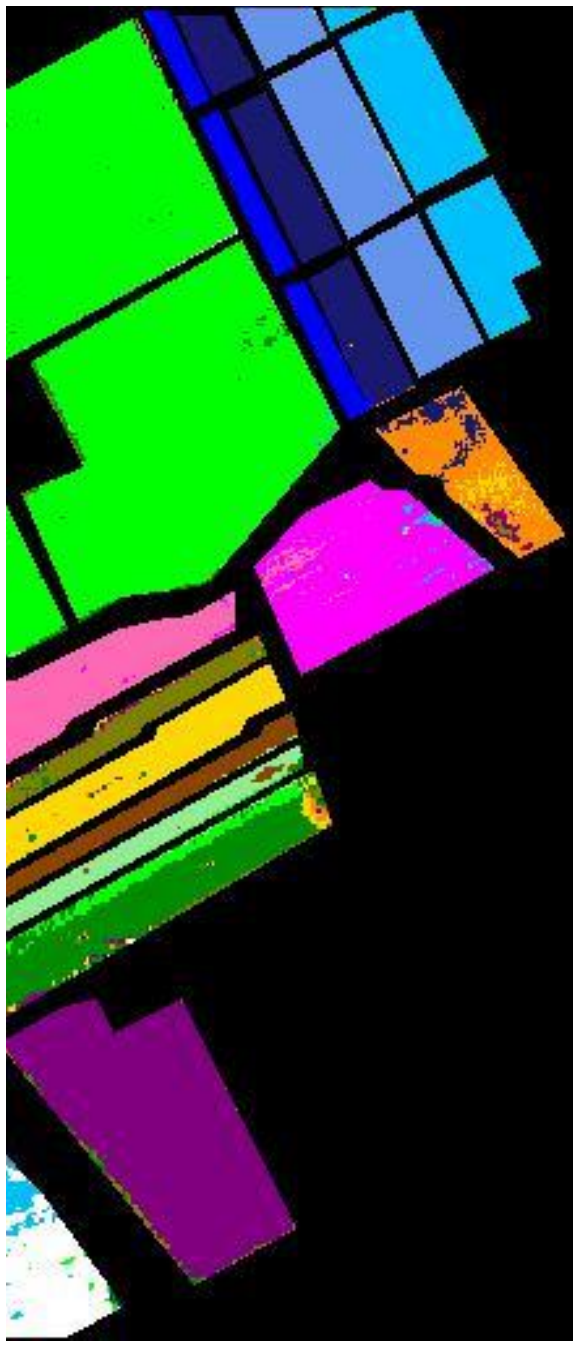}\label{fig:SVM}}
    \hspace{0.2cm}
    \subfigure[]{\includegraphics[width=0.18\textwidth]{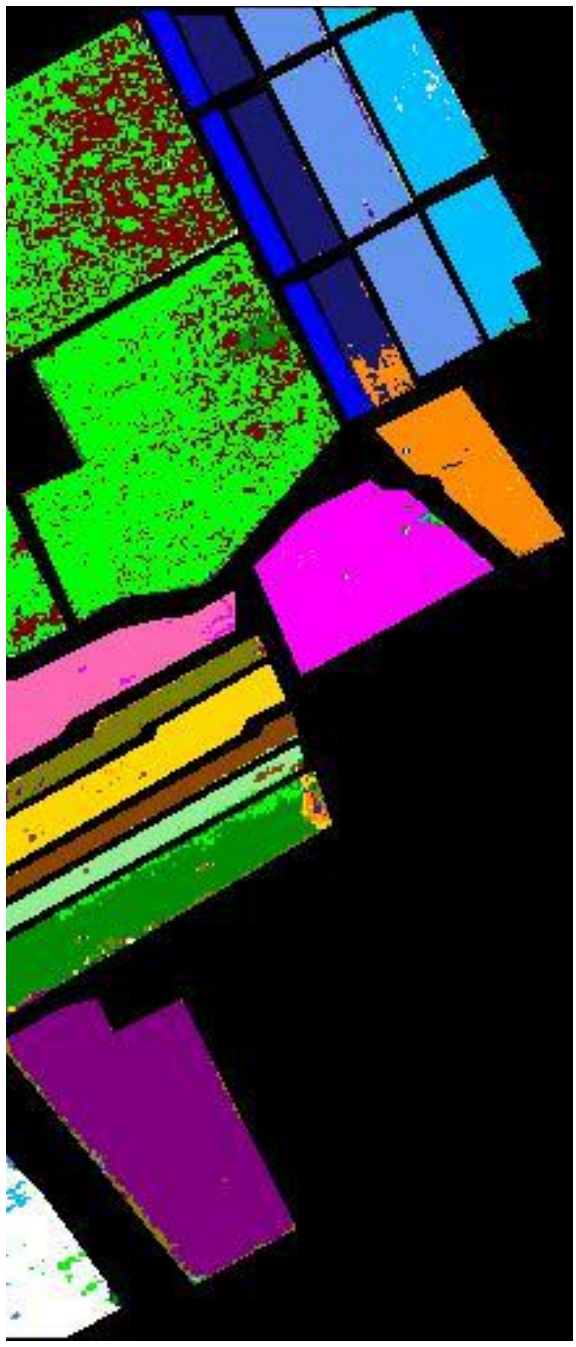}\label{fig:RNN}}

	\subfigure[]{\includegraphics[width=0.18\textwidth]{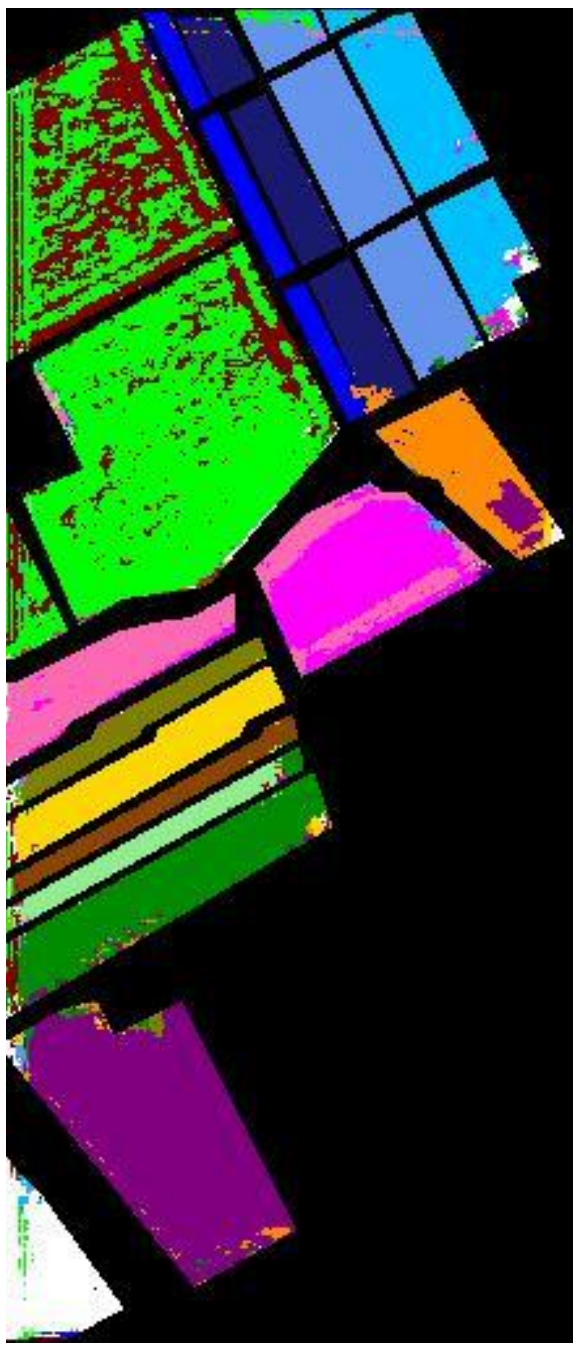}\label{fig:CNN}}
	\hspace{0.2cm}
    \subfigure[]{\includegraphics[width=0.18\textwidth]{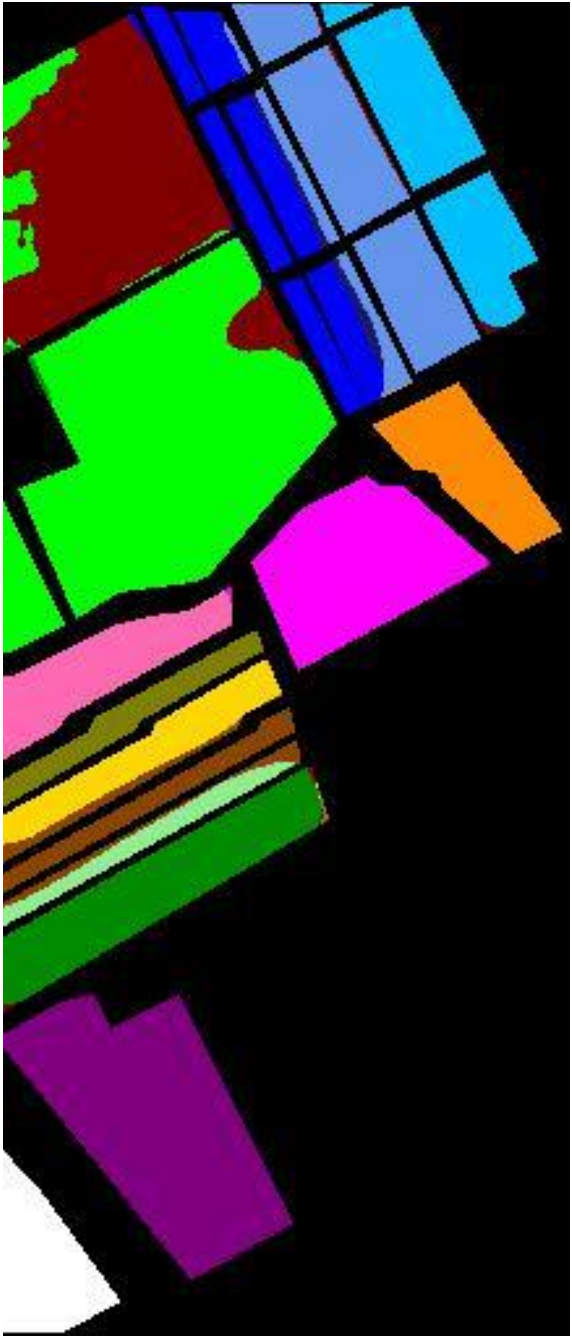}\label{fig:SRNN}}
	\hspace{0.2cm}
	\subfigure[]{\includegraphics[width=0.18\textwidth]{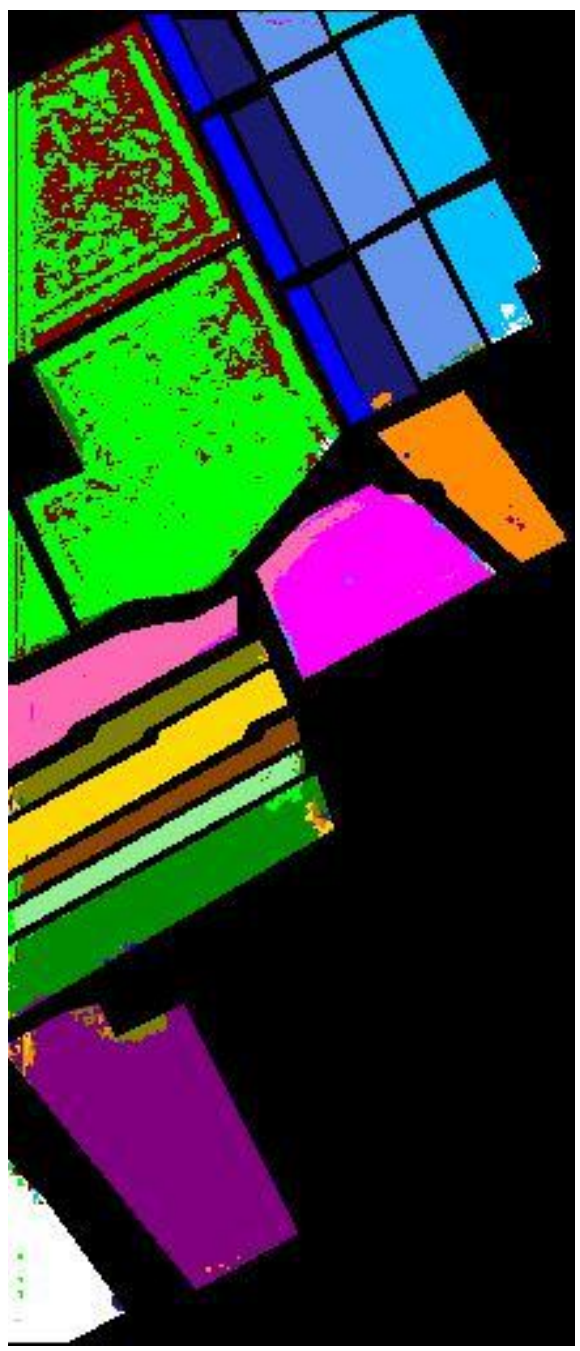}\label{fig:DCPE}}
	\hspace{0.2cm}
	\subfigure[]{\includegraphics[width=0.18\textwidth]{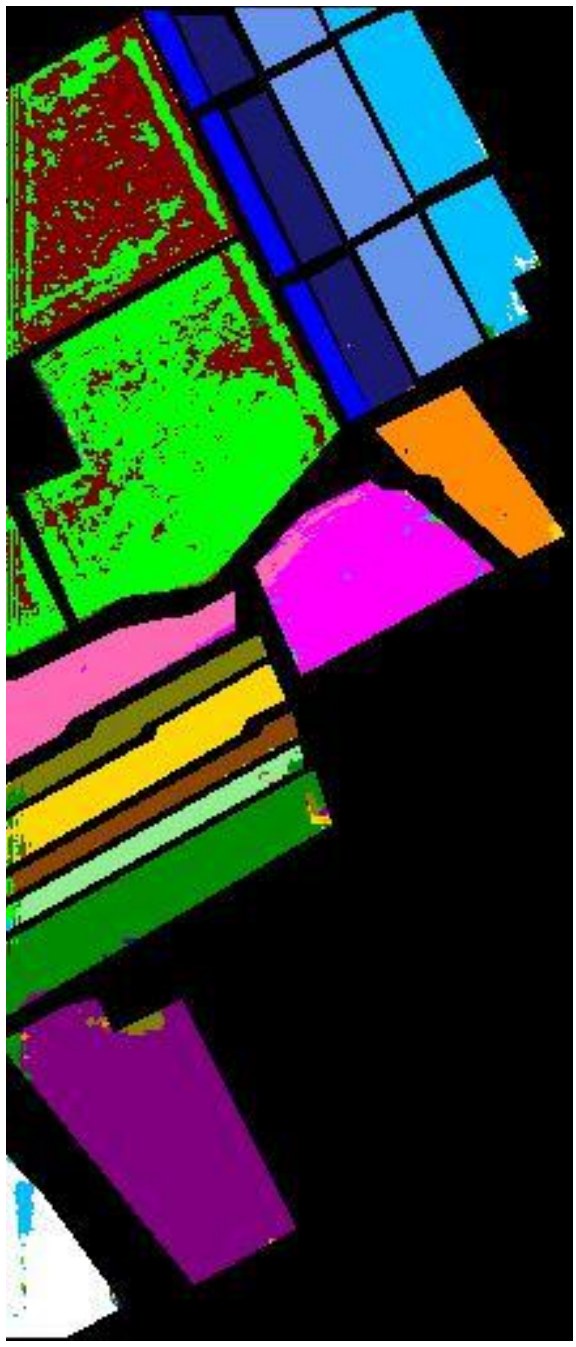}\label{fig:MDCPE}}
	\hspace{0.2cm}
    \subfigure{\includegraphics[width=0.18\textwidth]{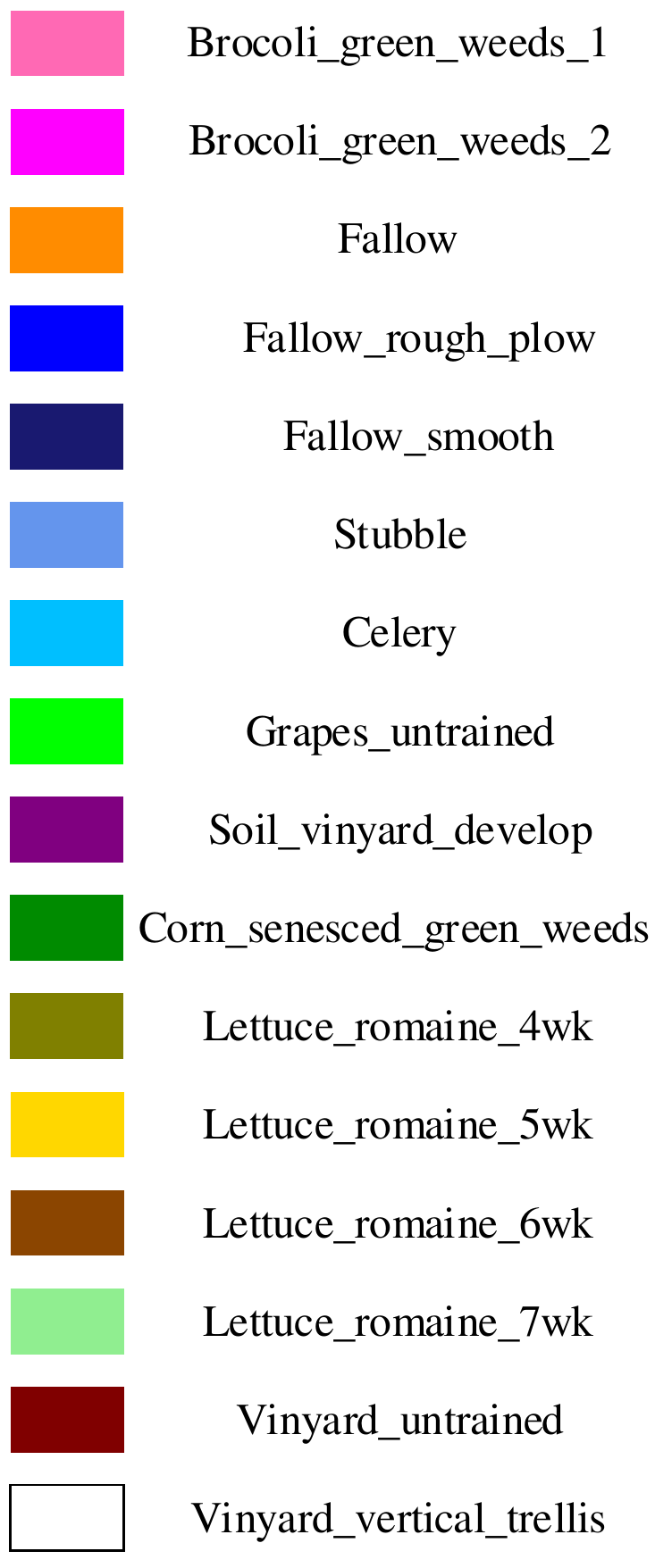}}
	\hspace{0.2cm}
	\caption{Classification results obtained by different methods for the Salinas scene. (a) False color image of HS data. (b) Ground-truth labels. (c) RF. (d) SVM. (e) RNN. (f) CNN. (g) SSRN. (h) DCPE-RNN-CNN. i) MDCPE-RNN-CNN.}\label{fig:SAFIGURE}		
\end{figure*}

\begin{enumerate}[1)]
\item RF: Random forest with 150 trees.
\item SVM: SVM with RBF kernel.
\item RNN: GRU-based recurrent neural network, which owns similar structure and configuration with the RNN-based learner used in MDCPE. Learning rate and the number of time steps are optimized on validation data.
\item CNN: Three dimensional convolutional neural network, which shares same structure and convolutional kernels with the CNN-based learner adopted in our method. Learning rate and spatial size of input block are optimized to fulfill the highest classification accuracy.
\item SSRN: An end-to-end spectral-spatial residual network proposed in \cite{zhong2018spectral}.
\item DCPE-RNN-CNN: RNN and CNN are taken as two learners of DCPE co-training. They share same structure with the learners in MDCPE.
\item MDCPE-RNN-CNN: The proposed co-training method based on MDCPE with RNN and CNN.
\end{enumerate}

To make a fair comparison, we utilize the same training data set and testing data set for all used methods and tuned these contrasts to their optimal settings in SA, PU and PC data sets. Overall accuracy (OA), average accuracy (AA) and Kappa coefficient (K) are all calculated as classification indexes during analyzing results.
\begin{figure*}[htbp]
	\centering
	\subfigure[]{\includegraphics[width=0.18\textwidth]{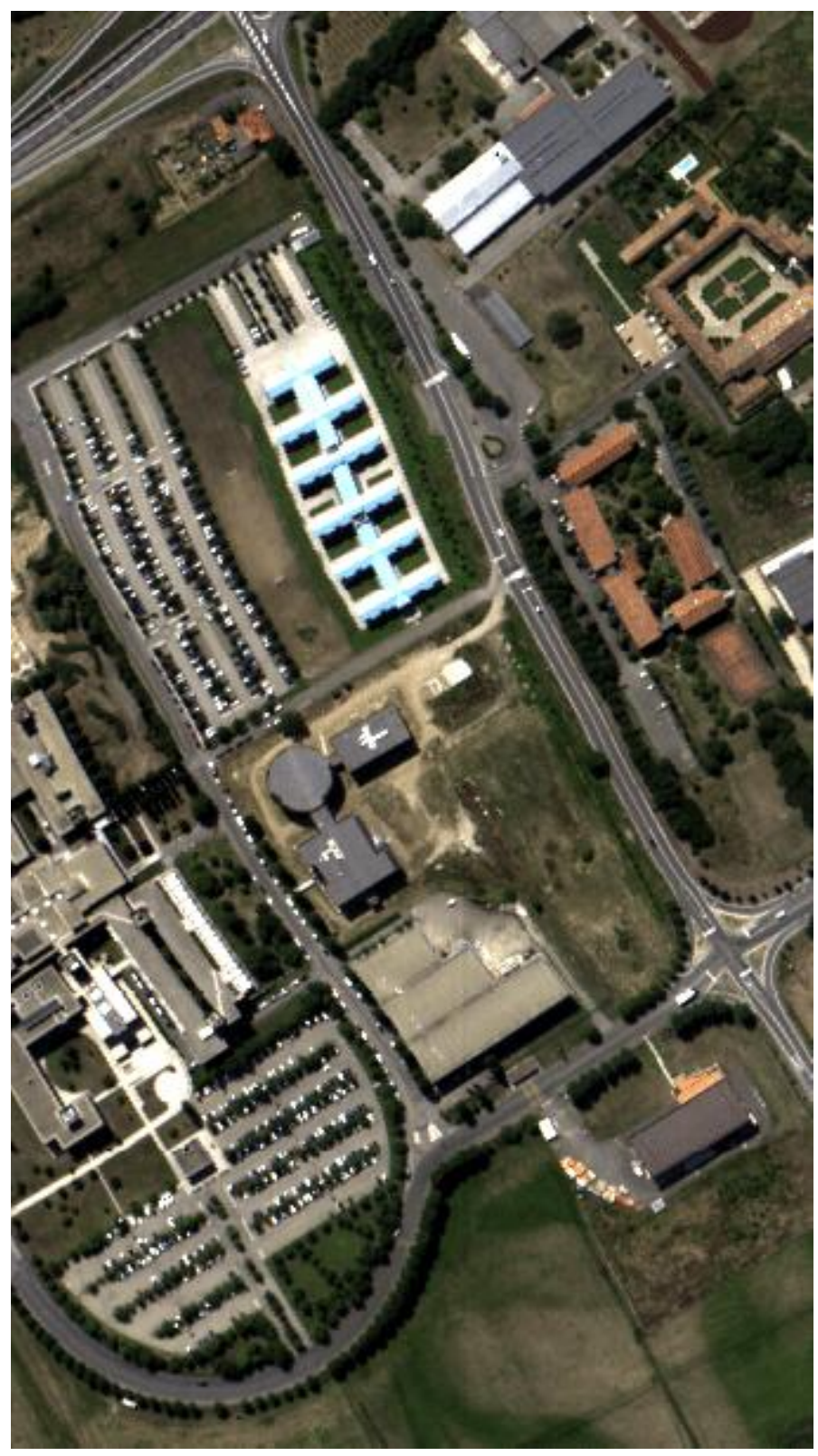}\label{fig:FCI}}
	\hspace{0.2cm}
	\subfigure[]{\includegraphics[width=0.18\textwidth]{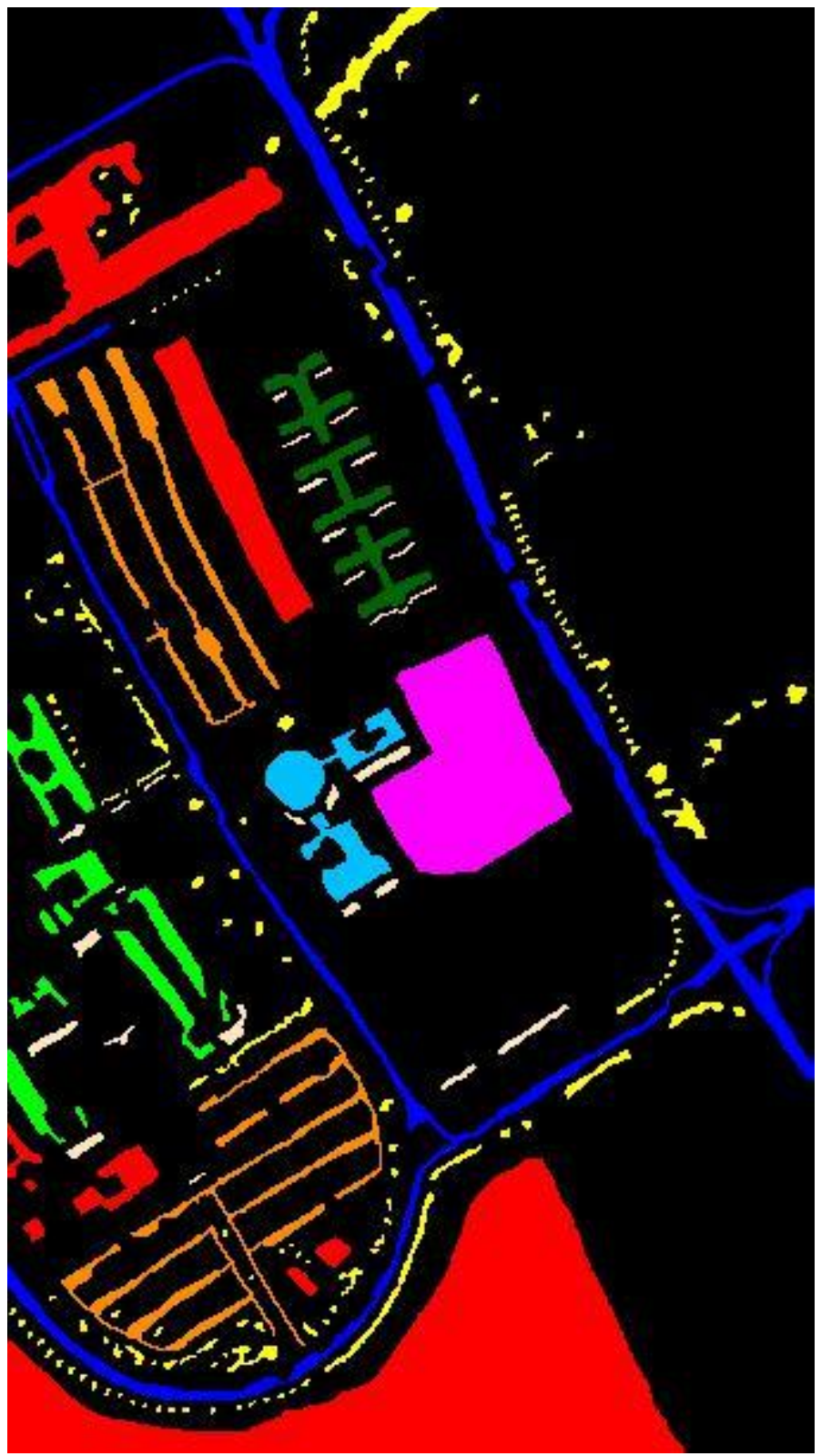}\label{fig:GT}}
	\hspace{0.2cm}
	\subfigure[]{\includegraphics[width=0.18\textwidth]{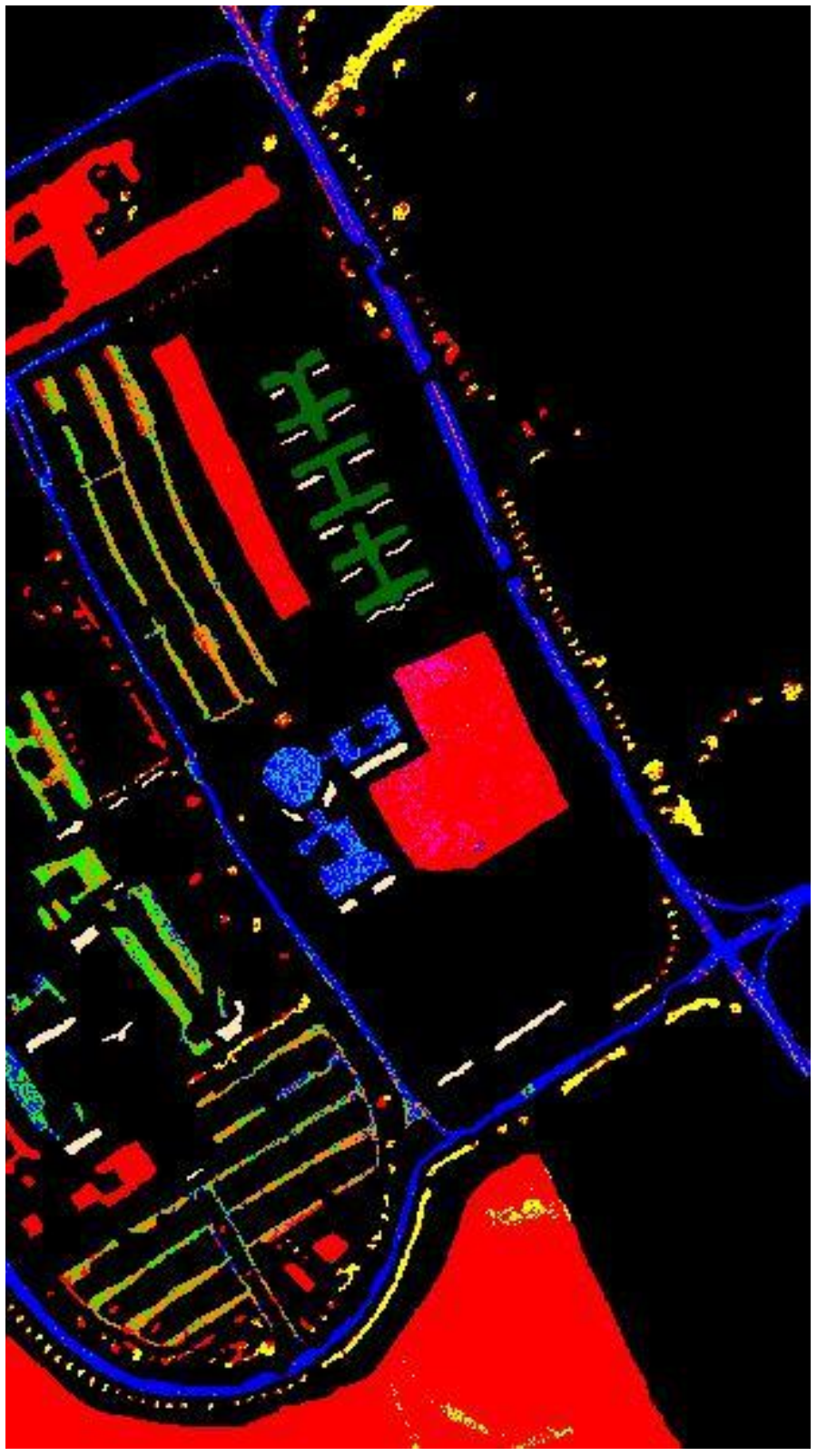}\label{fig:RF}}
	\hspace{0.2cm}
	\subfigure[]{\includegraphics[width=0.18\textwidth]{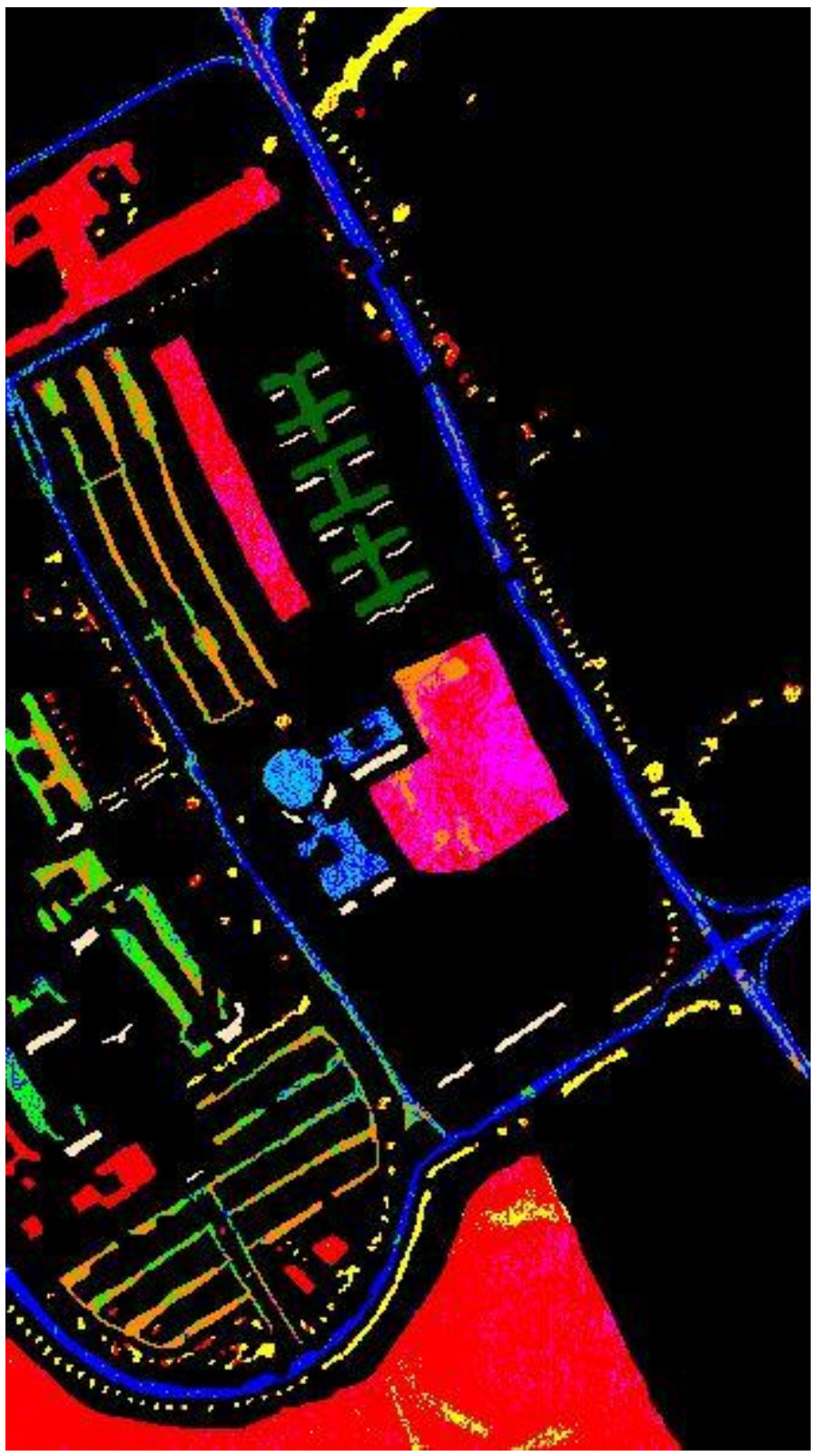}\label{fig:SVM}}
	\hspace{0.2cm}
    \subfigure[]{\includegraphics[width=0.18\textwidth]{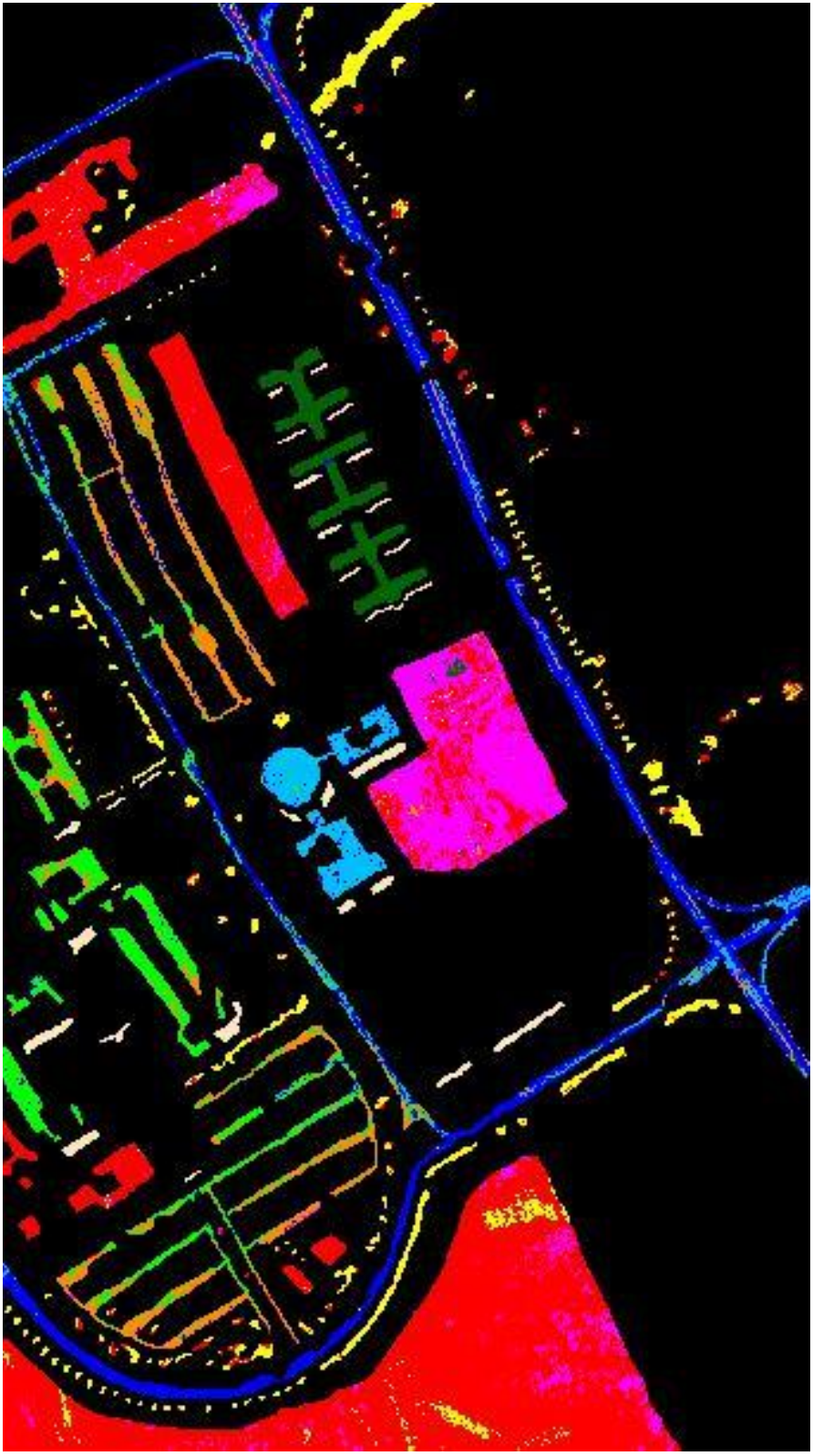}\label{fig:RNN}}

	\subfigure[]{\includegraphics[width=0.18\textwidth]{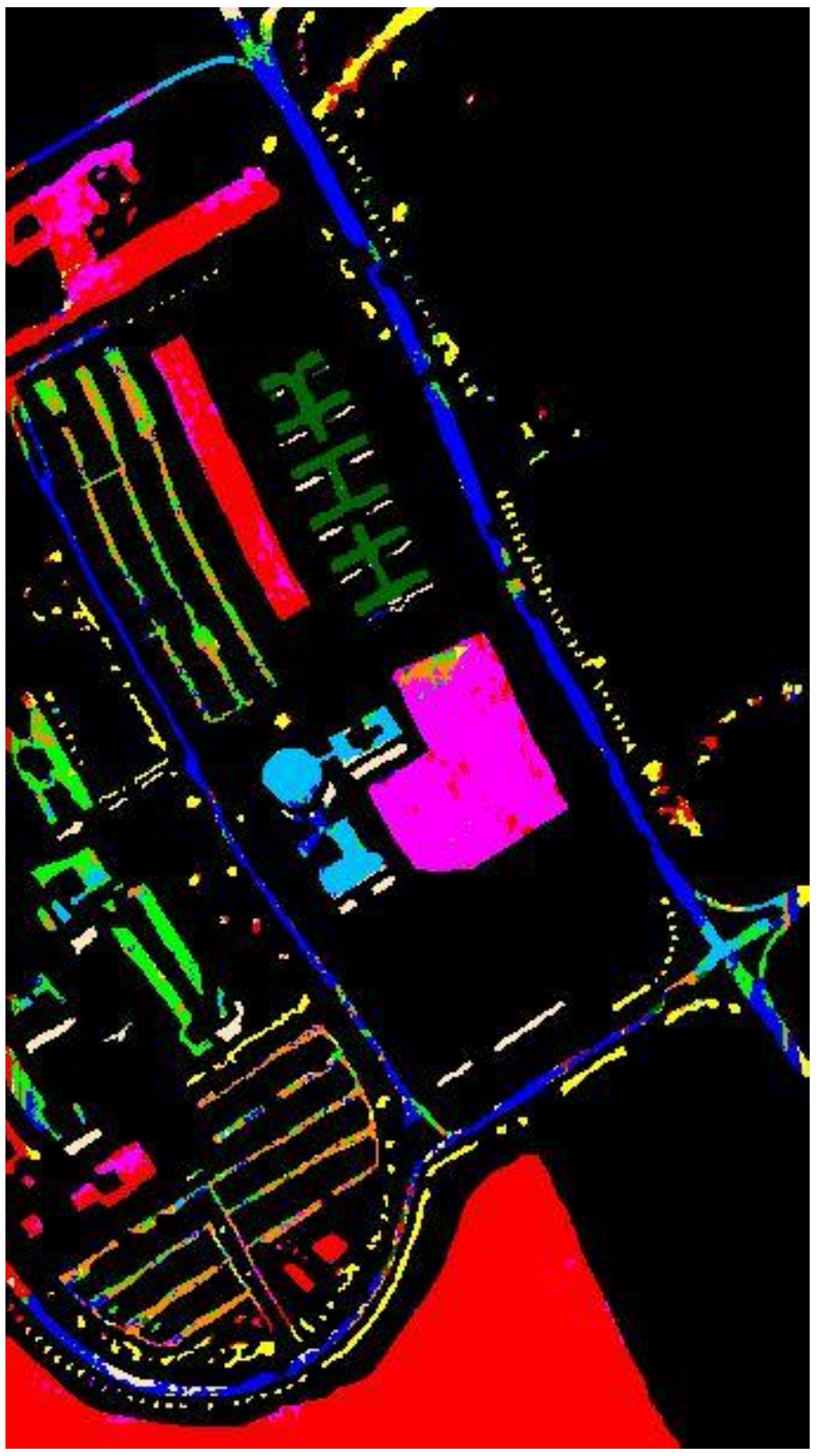}\label{fig:CNN}}
	\hspace{0.2cm}
    \subfigure[]{\includegraphics[width=0.18\textwidth]{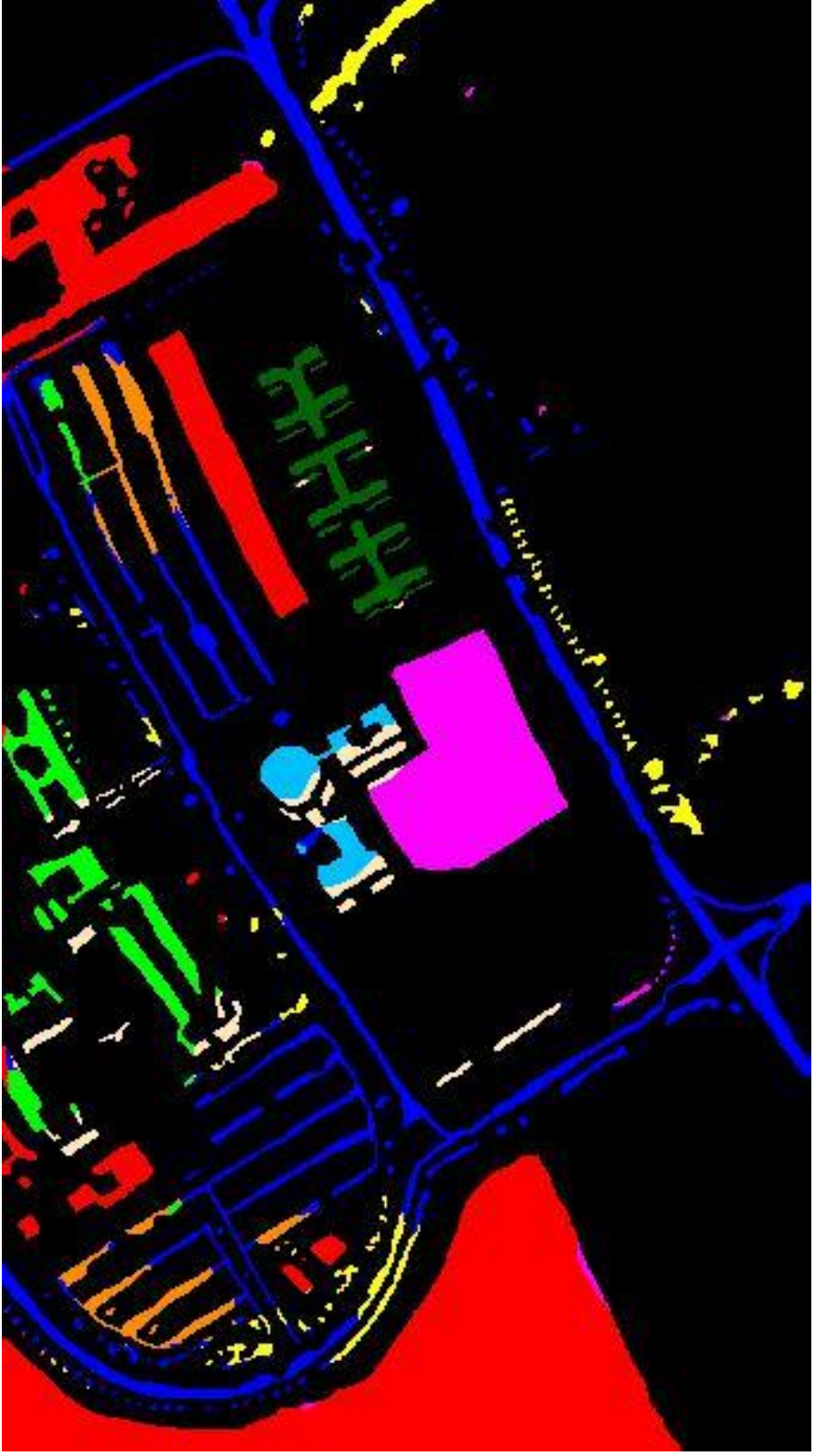}\label{fig:SSRN}}
	\hspace{0.2cm}
	\subfigure[]{\includegraphics[width=0.18\textwidth]{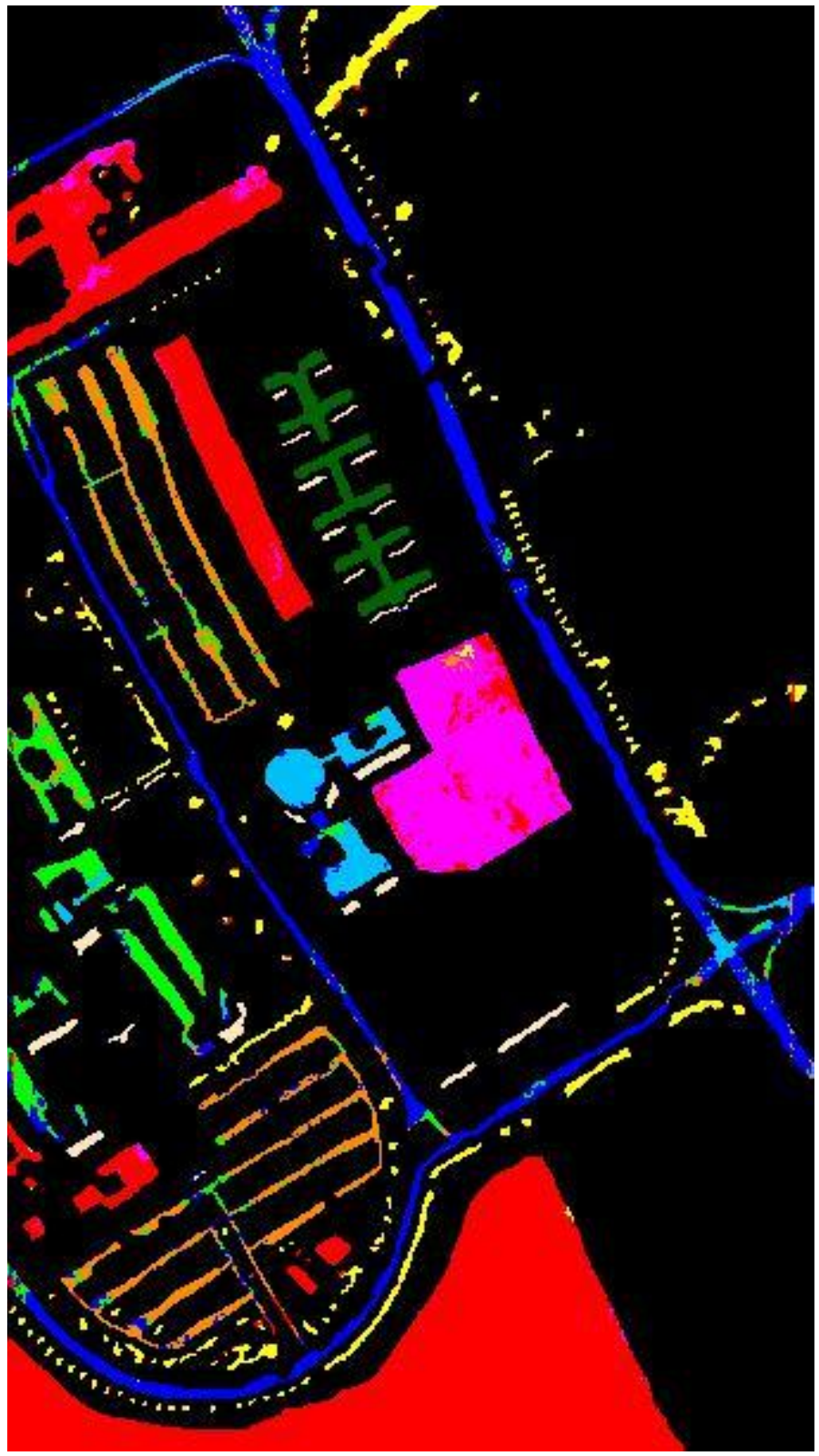}\label{fig:DCPE}}
	\hspace{0.2cm}
	\subfigure[]{\includegraphics[width=0.18\textwidth]{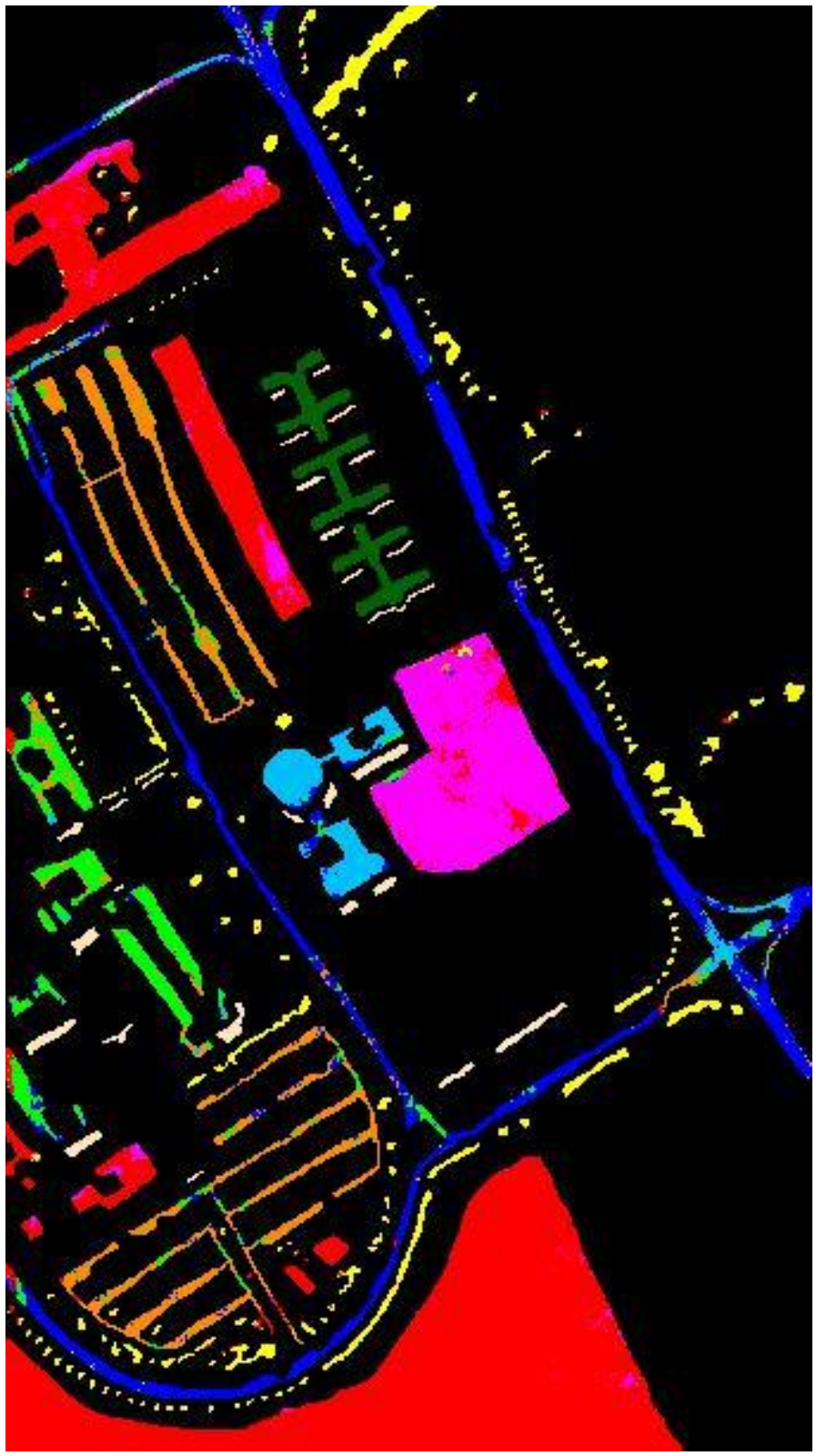}\label{fig:MDCPE}}
	\hspace{0.2cm}
    \subfigure{\includegraphics[width=0.18\textwidth]{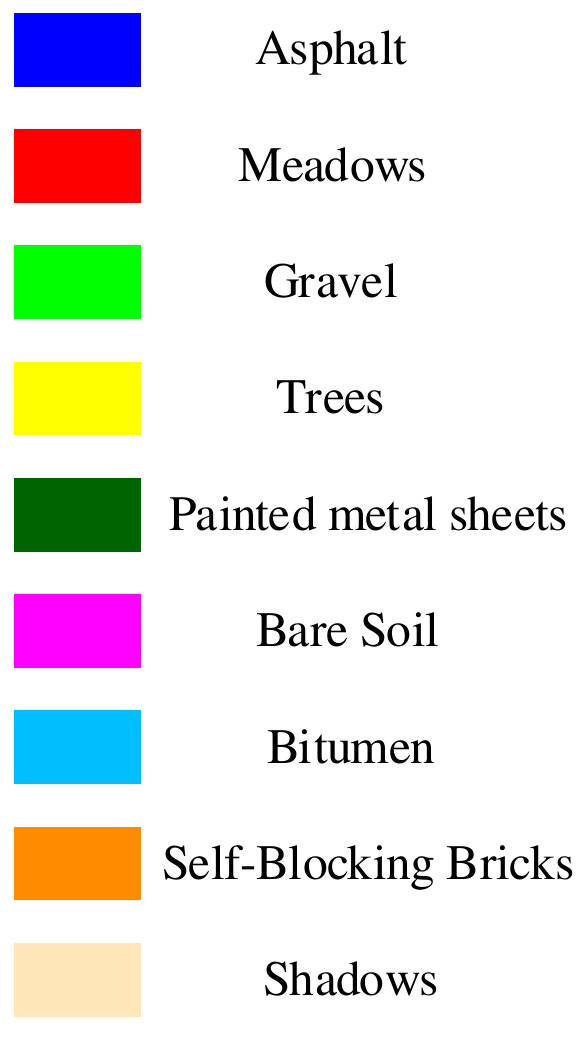}}
	\hspace{0.2cm}
	\caption{Classification results obtained by different methods for the Pavia University scene. (a) False color image of HS data. (b) Ground-truth labels. (c) RF. (d) SVM. (e) RNN. (f) CNN. (g) SSRN. (h) DCPE-RNN-CNN. i) MDCPE-RNN-CNN.}\label{fig:PUFIGURE}		
\end{figure*}

\subsubsection{Results on SA Data Set}
Table \ref{tab:SA_acc} shows the classification indexes on SA data set, including OAs, AAs, Kappa coefficients, and the classification accuracy of each class. It's clear that SSRN and the proposed method have better performance than others. Traditional random tree (RF) and support vector machine (SVM) merely take advantage of the shallow spectral feature but ignoring abundant spatial feature, which leads to modest classification performance. However, SVM has the best performance on class 1, 8 and 13. RNN and CNN are utilized as learners to extract spectral or spatial feature separately. All of their performance are unfavorable because only spectral or spatial feature is insufficient for HSI classification. Comparing with DCPE-RNN-CNN, the prominent advantage of proposed method is that it enhances the accuracy of indistinguishable classes. For instance, table \ref{tab:SA_acc} shows that the accuracy of MDCPE-RNN-CNN on class 15 is 72.94\%, which is far higher than that of DCPE-RNN-CNN. After improving the performance on the classes that are hard to differentiate, our method increases the accuracy by 2.45\% of OA, 1.25\% of AA and 0.0277 of Kappa coefficient compared with DCPE-RNN-CNN. SSRN has the best accuracy on several classes and thus slightly breach the proposed method on OA and Kappa coefficient. However, MDCPE-RNN-CNN surpass a lot than SSRN on class 4 and 13, reflecting that the proposed method is more homogeneous and robust, not only achieves good performance in large area but also identifies small and indistinctive objects.

The classification maps of the best trained models on SA data set are visualized in Figs. \ref{fig:SAFIGURE}. The false color images of original HSI data and their corresponding ground-truth maps are also shown along side with classification maps to offer direct comparison. Some classes that are hard to be correctly classified challenges the robustness and effectiveness of classifier. One possible reason is that these indistinguishable classes have similar spectral and spatial feature with other classes due to their complex and confusing ground-truth. In Fig. \ref{fig:SAFIGURE}, All methods mistake the area of class Vinyard\_untrained for class Grapes\_untrained, as these two classes are mix up and hard to classify in true surfaces. Although further improvement is needed for SSRN and MDCPE-RNN-CNN in this class, more pixels are classified correctly and most similar classification map with ground-truth labels is obtained by this two methods. Classification map of SSRN seems more homogeneous and smooth than other maps in most classes, but it has poor performance on class Fallow\_rough\_plow and Lettuce\_romaine\_6wk.
\begin{table*}[htbp]
\centering
\caption{CLASSIFICATION RESULTS OF DIFFERENT METHODS FOR THE PC DATA SET. THE BEST ACCURACY IN EACH ROW IS SHOWN IN BOLD}
\label{tab:PC_acc}
\includegraphics[width=0.95\textwidth]{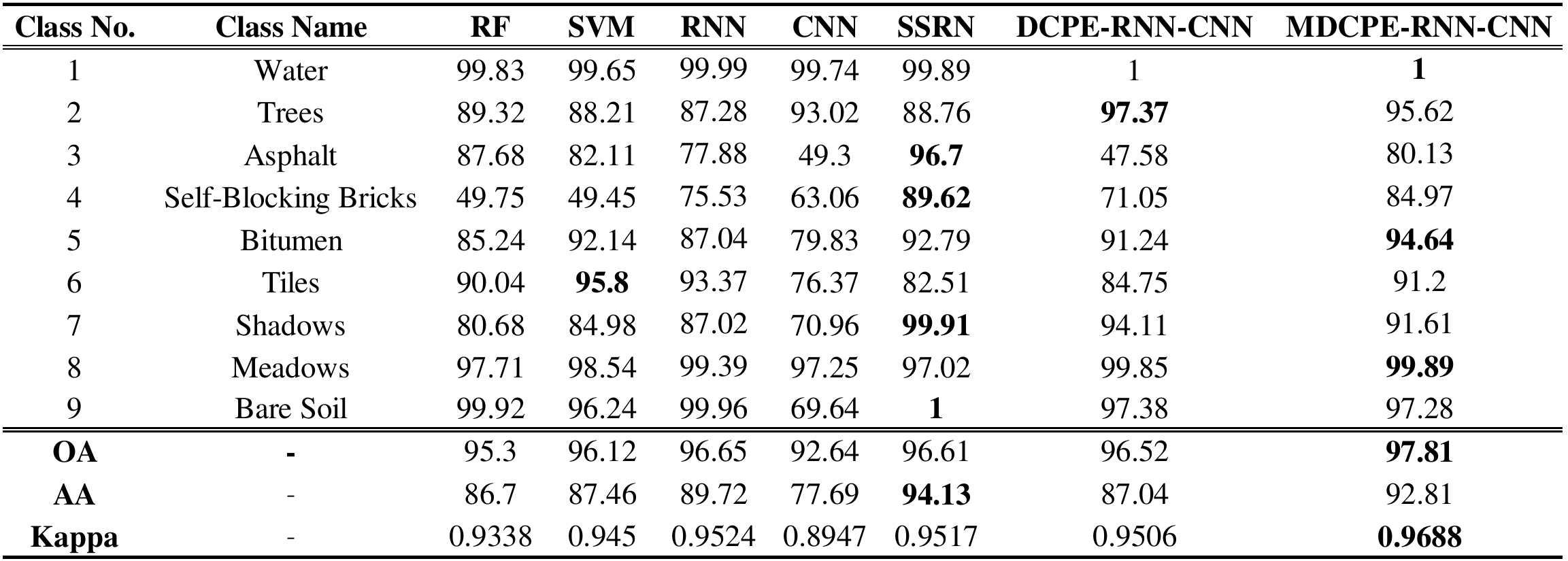}
\end{table*}
\subsubsection{Results on PU Data Set}
Quantitative classification results of PU data set are shown in Table \ref{tab:PU_acc}. It's obvious that the proposed MDCPE-RNN-CNN, DCPE-RNN-CNN and SSRN surpass others. SSRN owns highest accuracy on class 3, 6, 7, and 8 but it has poor performance on class 1 and 9. Such unbalanced classification accuracies between classes impairs overall effectiveness of the method. DCPE-RNN-CNN and MDCPE-RNN-CNN have better performance than others because 1) these two co-training methods resort to two deep neural networks(RNN, CNN) as two learners to extract more discriminative feature from spectral and spatial views, 2) rich and valuable unlabeled data are adopted to enhance the generality and robustness of learners. DCPE-RNN-CNN has low precision on class 3 and 8 even it acquires highest accuracy on class 5. The proposed MDCPE-RNN-CNN manifests the highest accuracy and well-balanced accuracy between classes. Comparing to SSRN and DCPE-RNN-CNN, MDCPE-RNN-CNN increases the accuracy significantly by 2.37, 2.7\% of OA, 1.37, 3.09\% of AA and 0.0332, 0.0354 of Kappa coefficient, respectively.

Fig. \ref{fig:PUFIGURE} shows the classification maps of the PU data set qualitatively. The proposed method achieves the better performance on class Trees, Painted Metal Sheets and Shadows. These classes occupy either large areas of corridor or piecemeal mapped to ground-truths, for which they are easily perturbed by noise and other neighboring classes during the process of spectral imaging. Another reason is that the classifiers fail to learn differentiable feature of such classes due to few and unbalanced labeled training data. Although it's hard to learn the features from such classes and to classify them correctly, our method surpass other approaches in acquiring more homogeneous and well-bounded classification maps.
\begin{figure*}[htbp]
	\centering
	\subfigure[]{\includegraphics[width=0.18\textwidth]{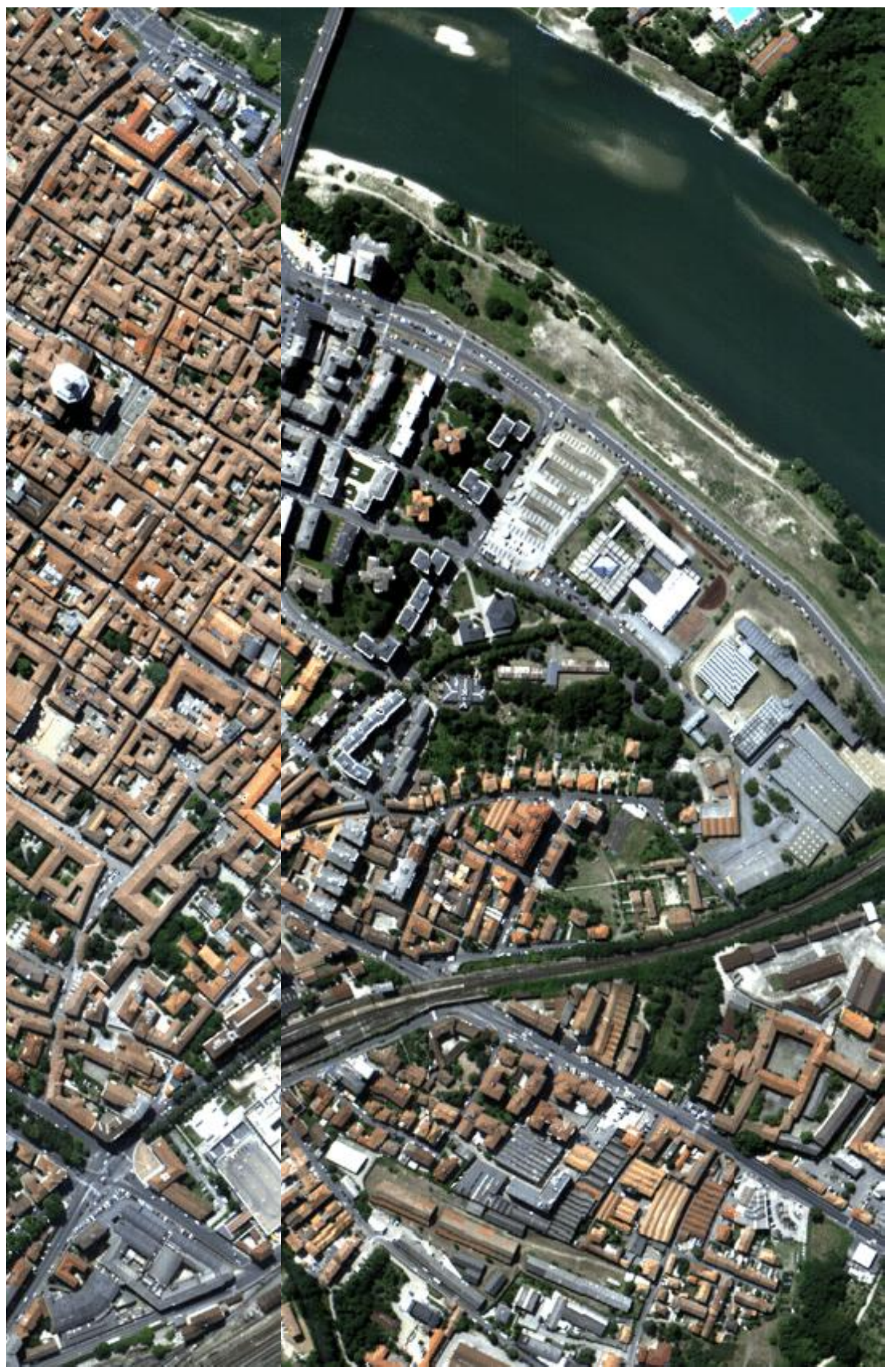}\label{fig:FCI}}
	\hspace{0.2cm}
	\subfigure[]{\includegraphics[width=0.18\textwidth]{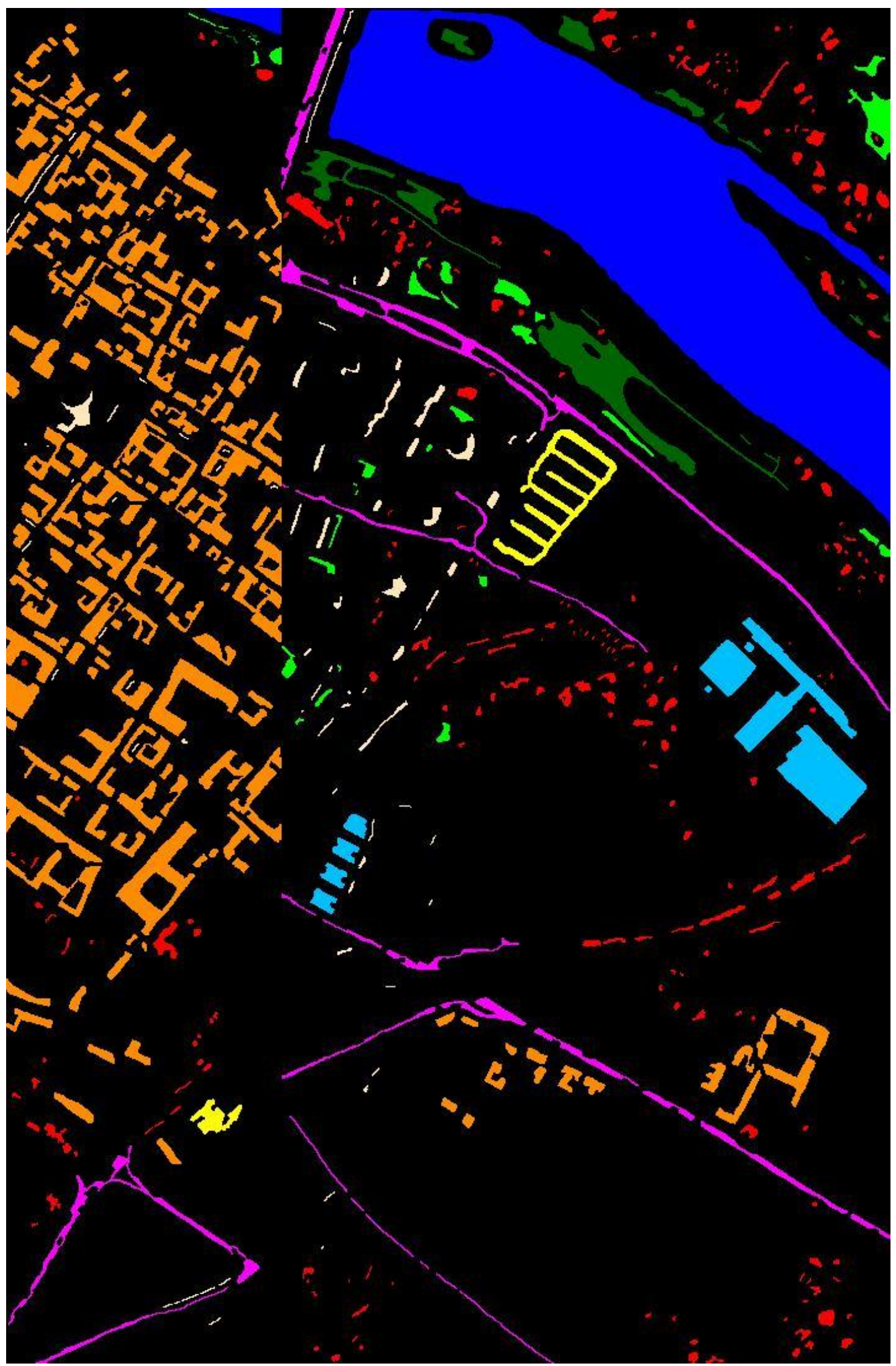}\label{fig:GT}}
	\hspace{0.2cm}
	\subfigure[]{\includegraphics[width=0.18\textwidth]{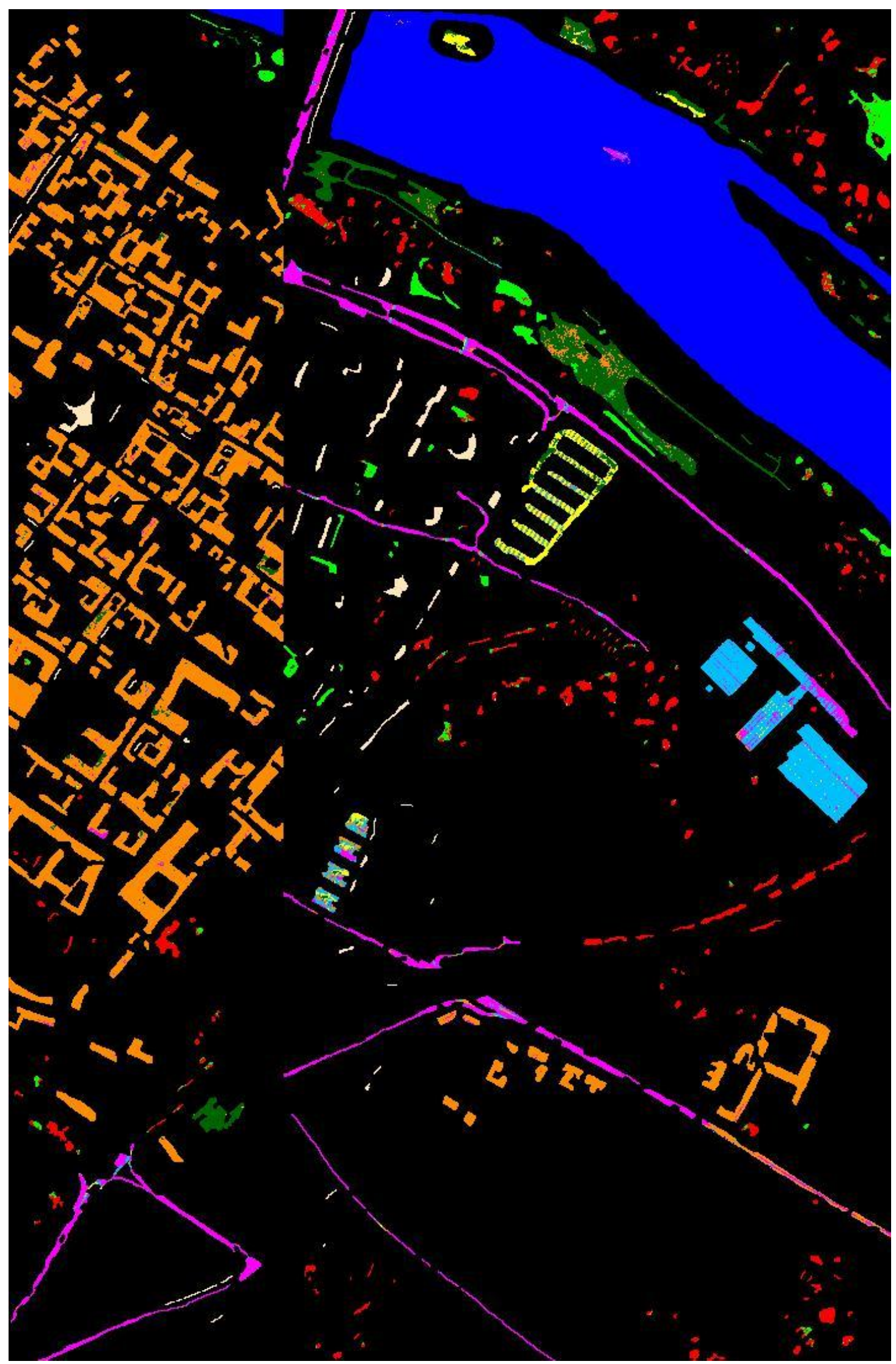}\label{fig:RF}}
	\hspace{0.2cm}
	\subfigure[]{\includegraphics[width=0.18\textwidth]{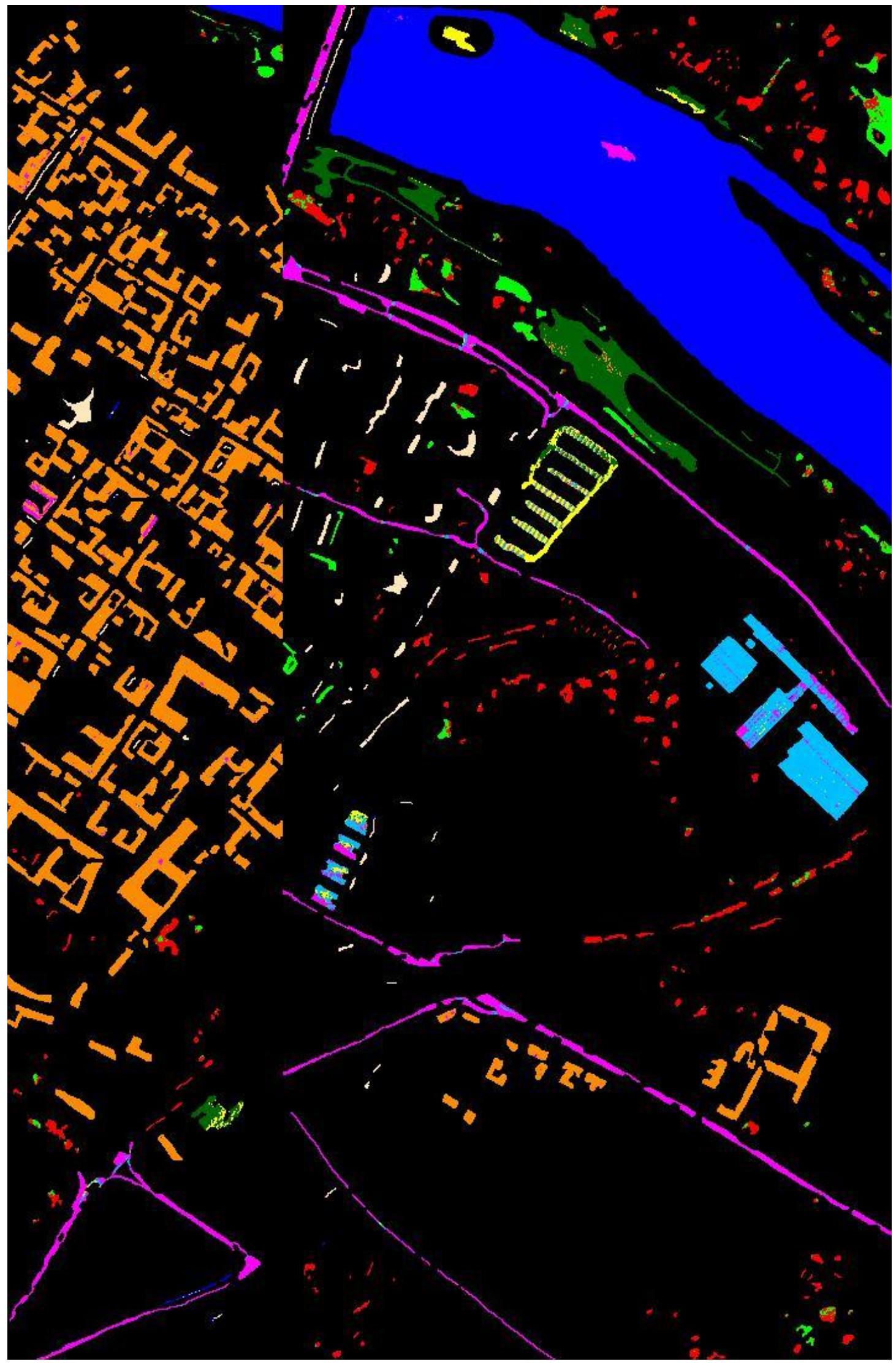}\label{fig:SVM}}
	\hspace{0.2cm}
    \subfigure[]{\includegraphics[width=0.18\textwidth]{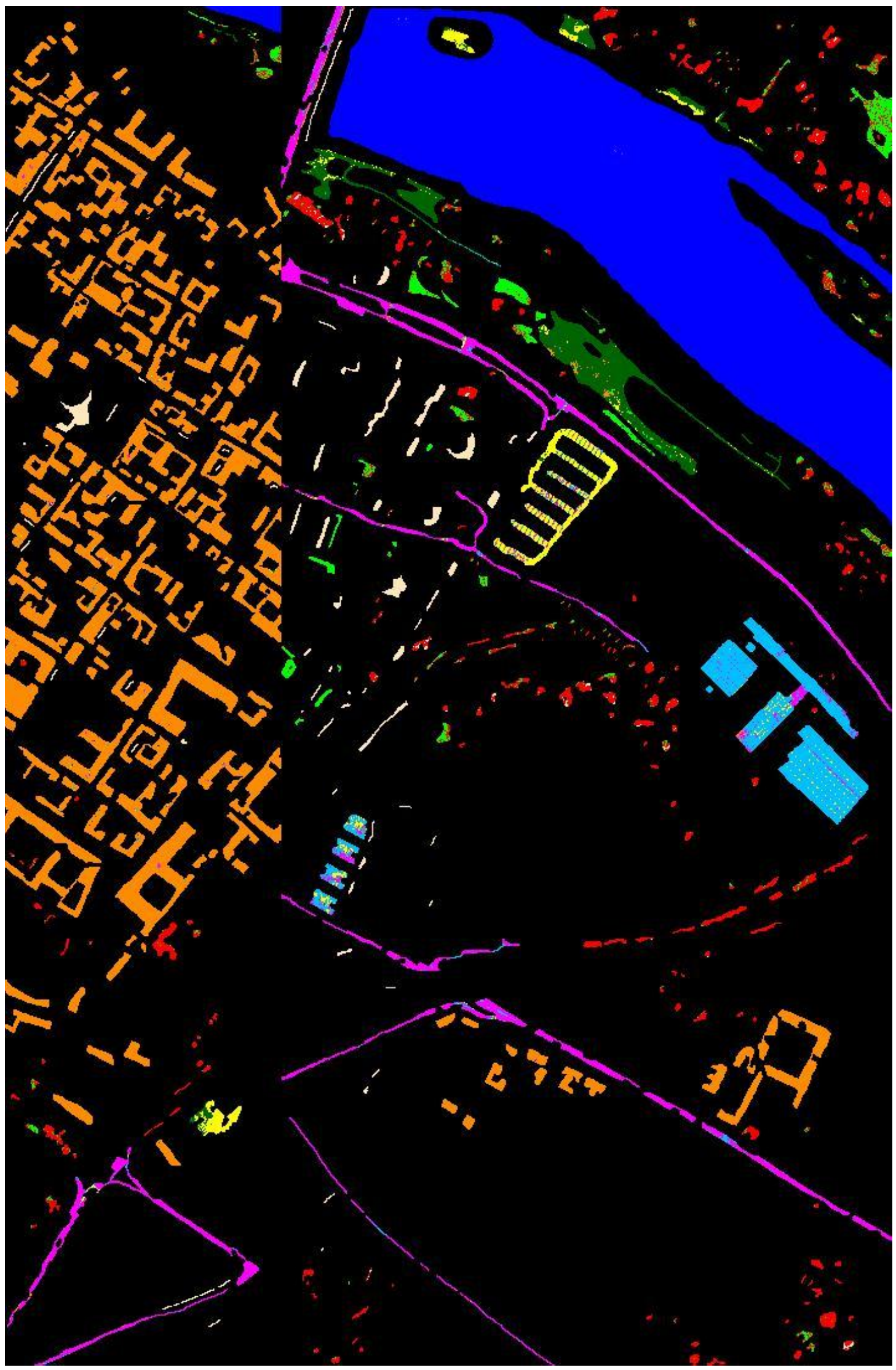}\label{fig:RNN}}
	
	\subfigure[]{\includegraphics[width=0.18\textwidth]{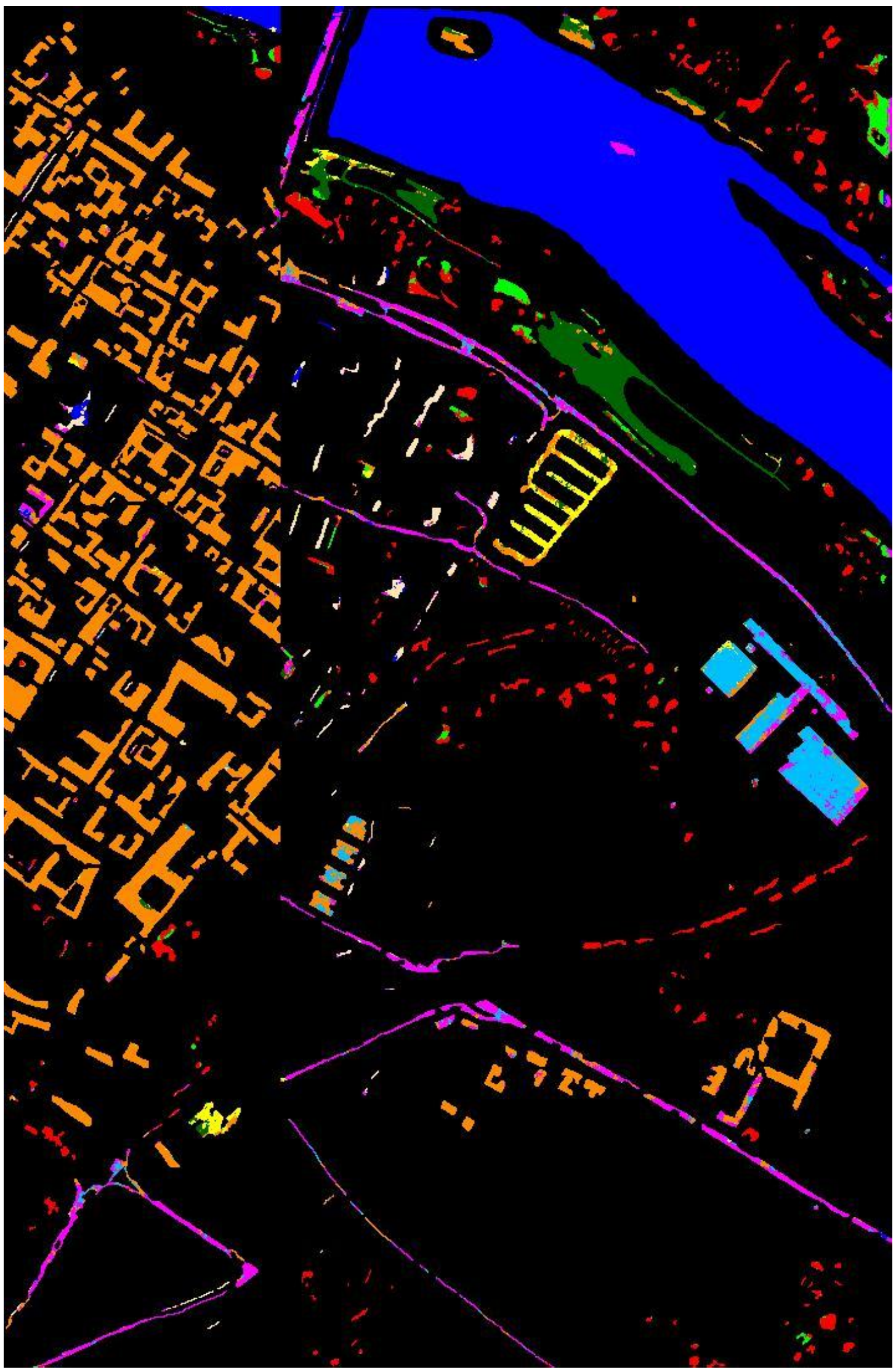}\label{fig:CNN}}
	\hspace{0.2cm}
    \subfigure[]{\includegraphics[width=0.18\textwidth]{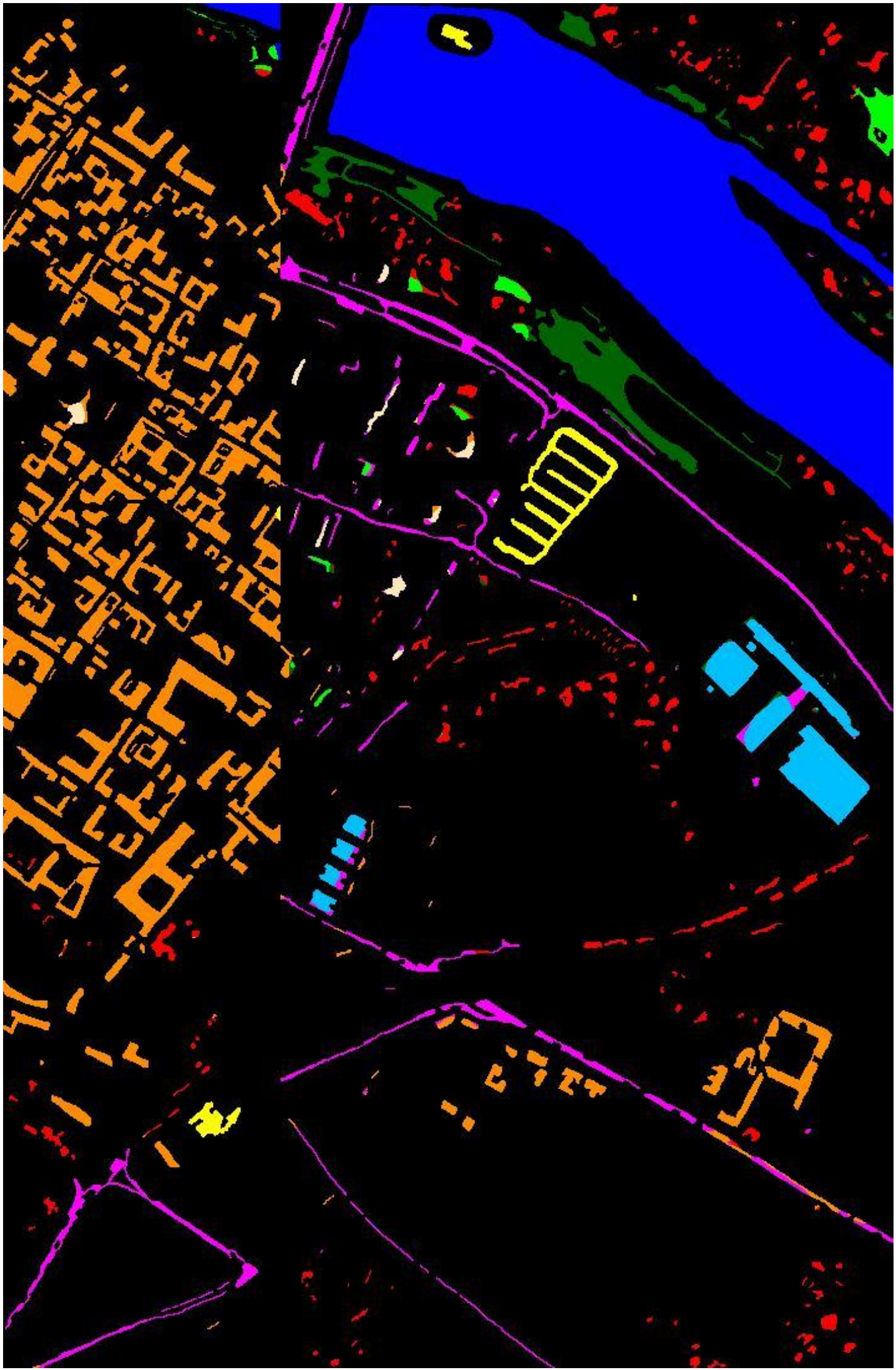}\label{fig:SSRN}}
	\hspace{0.2cm}
	\subfigure[]{\includegraphics[width=0.18\textwidth]{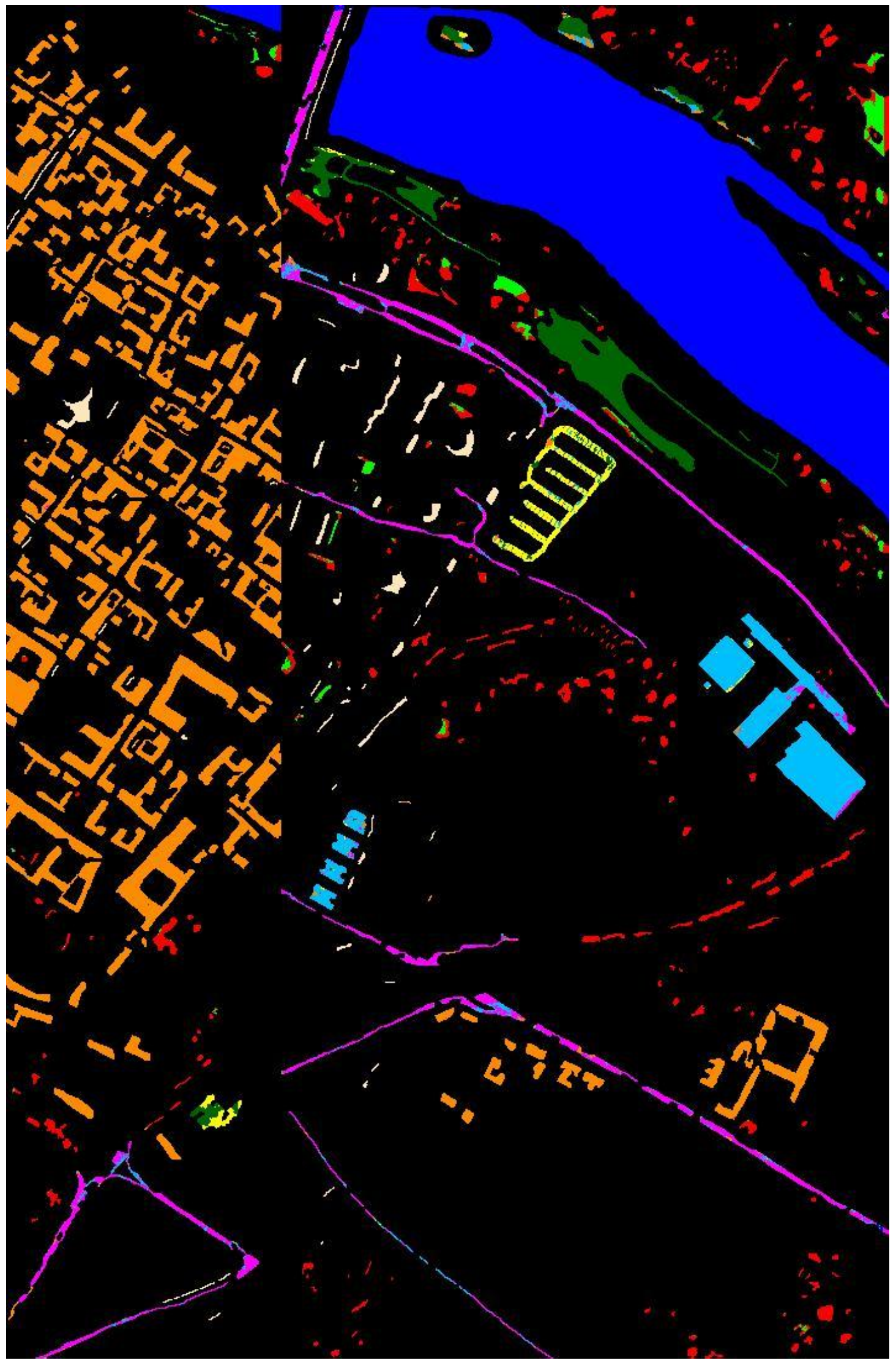}\label{fig:DCPE}}
	\hspace{0.2cm}
	\subfigure[]{\includegraphics[width=0.18\textwidth]{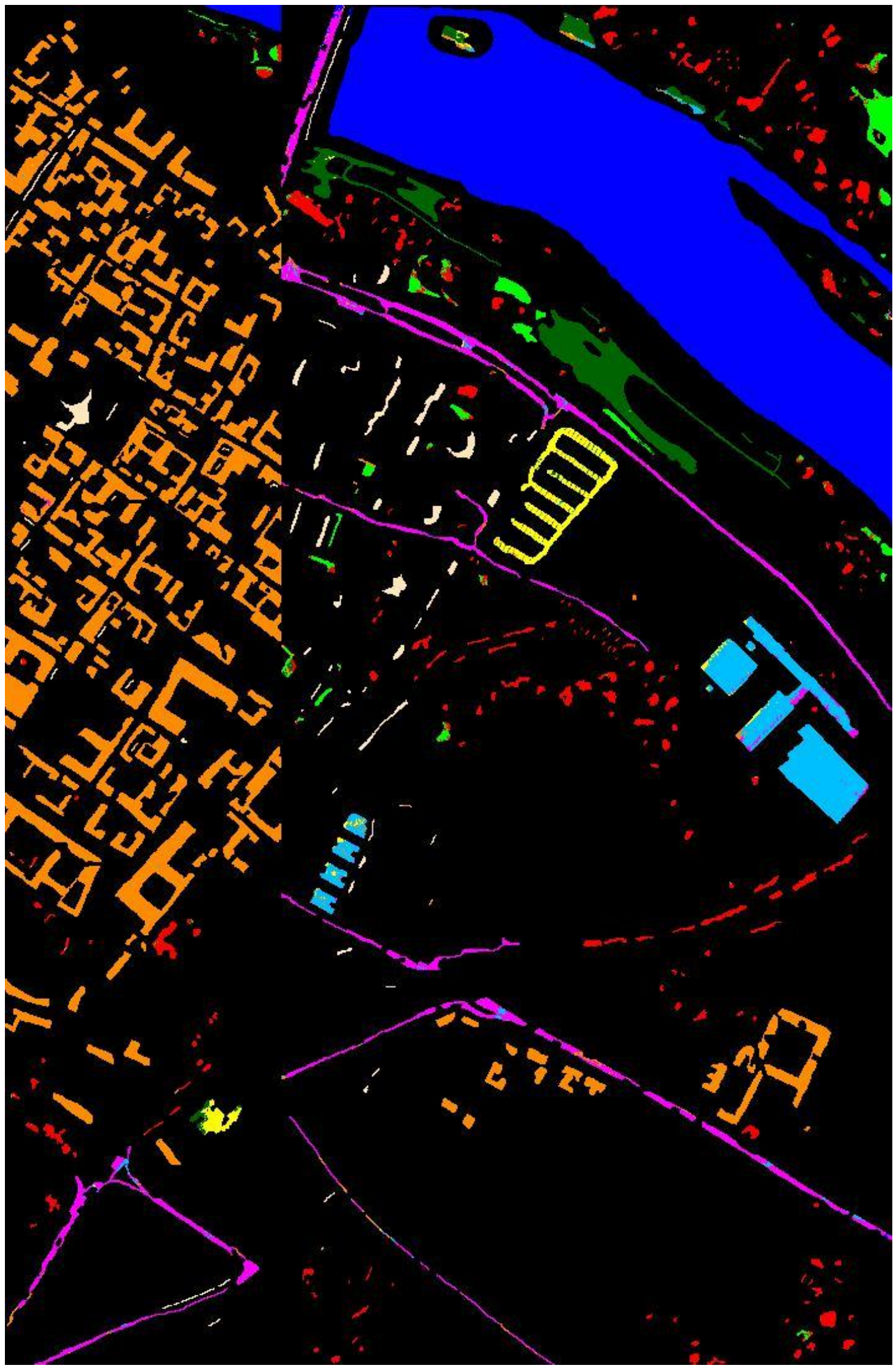}\label{fig:MDCPE}}
	\hspace{0.2cm}
    \subfigure{\includegraphics[width=0.18\textwidth]{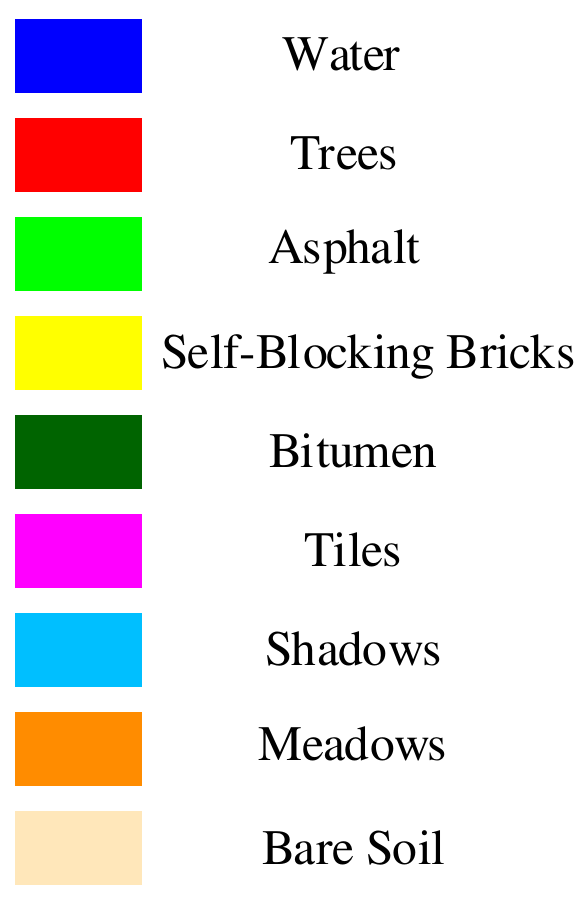}}
	\hspace{0.2cm}
	\caption{Classification results obtained by different methods for the Pavia Center scene. (a) False color image of HS data. (b) Ground-truth labels. (c) RF. (d) SVM. (e) RNN. (f) CNN. (g) SSRN. (h) DCPE-RNN-CNN. i) MDCPE-RNN-CNN.}\label{fig:PCFIGURE}			
\end{figure*}

\subsubsection{Results on PC Data Set}
The classification maps of the PC data set obtained by the vector-based models, deep neural networks and our method are shown in Fig. \ref{fig:PCFIGURE}, and the corresponding accuracy indexes are presented in Table \ref{tab:PC_acc}. Kappa is thought to be a more persuasive index than a simple percent agreement calculation, since it considers the agreement occurring by chance [46]. As table \ref{tab:PC_acc} shows, MDCPE-RNN-CNN generates highest Kappa and the best classification results. RF, SVM, and RNN have very low accuracy on class 4, and CNN, DCPE-RNN-CNN shows low-precision on class 3 and 4. Although SSRN acquires highest AA among other methods, its OA and Kappa coefficients are 1.2 percent and 0.0171 lower than the proposed method.

As Fig. \ref{fig:PCFIGURE} shows, classification maps of other methods present many noisy points and confuse class Self-Blocking Bricks with class Meadows and Bitumen. But the proposed method and SSRN acquire more smooth and homogeneous results on class Self-Blocking Bricks and Meadows. Moreover, Results on class Shadows of other methods are messy and show severe misclassification, but the proposed method display more accurate and uniform performance on this class. Taking accuracies of all classes into consideration, our method shows more robust results even though small amount of samples and unbalance between classes.

\section{Conclusions}

In this paper, a semi-supervised co-training method is proposed to extract spectral-spatial feature based on MDCPE of RNN and CNN for HSI classification. RNN and CNN are utilized as two learners of MDCPE co-training to extract spectral and spatial features, respectively. MDCPE is then adopted to co-train two learners based on extracted spectral-spatial feature, which is conducive to keep updated data balanced between classes and promotes the generality of learners with unlabeled data. Analyzing experiment results on three widely studied HSI cases, our method not only acquires higher and more balanced accuracy than other contrast methods on the quantitative, but also presents more homogeneous and refined classification maps in visual. Moreover, experiments results prove that balanced training data and unlabeled data have great potential in improving the classification accuracy, which motivates us to make full use of unlabeled data and ensure the balance across classes in further study. Considering the favorable performance with limited and imbalanced labeled data, we believe that the proposed method can outperform other methods in more of HSI classification cases.

Even though the proposed MDCPE co-training surpasses comparative methods as shown in experiment results, further study is deserved to promote performance on the classes that are indiscriminating and can not be right classified in MDCPE. Besides, two neural networks are adopted as learners of MDCPE co-training to extract spectral and spatial features separately, which is considerably complicated and time-consuming. A intergraded learner is expected to come out, which can extract spectral and spatial features simultaneously and use unlabeled data to enhance the classification performance when labeled data is inadequate and imbalanced.

\ifCLASSOPTIONcaptionsoff
  \newpage
\fi

\bibliographystyle{IEEEtran}
\bibliography{myreferences}

\begin{thebibliography}{10}
\providecommand{\url}[1]{#1}
\csname url@samestyle\endcsname
\providecommand{\newblock}{\relax}
\providecommand{\bibinfo}[2]{#2}
\providecommand{\BIBentrySTDinterwordspacing}{\spaceskip=0pt\relax}
\providecommand{\BIBentryALTinterwordstretchfactor}{4}
\providecommand{\BIBentryALTinterwordspacing}{\spaceskip=\fontdimen2\font plus
\BIBentryALTinterwordstretchfactor\fontdimen3\font minus
  \fontdimen4\font\relax}
\providecommand{\BIBforeignlanguage}[2]{{%
\expandafter\ifx\csname l@#1\endcsname\relax
\typeout{** WARNING: IEEEtran.bst: No hyphenation pattern has been}%
\typeout{** loaded for the language `#1'. Using the pattern for}%
\typeout{** the default language instead.}%
\else
\language=\csname l@#1\endcsname
\fi
#2}}
\providecommand{\BIBdecl}{\relax}
\BIBdecl

\bibitem{gu2016nonlinear}
Y.~Gu, T.~Liu, X.~Jia, J.~A. Benediktsson, and J.~Chanussot, ``Nonlinear
  multiple kernel learning with multiple-structure-element extended
  morphological profiles for hyperspectral image classification,'' \emph{IEEE
  Transactions on Geoscience and Remote Sensing}, vol.~54, no.~6, pp.
  3235--3247, 2016.

\bibitem{li2016discontinuity}
J.~Li, M.~Khodadadzadeh, A.~Plaza, X.~Jia, and J.~M. Bioucas-Dias, ``A
  discontinuity preserving relaxation scheme for spectral--spatial
  hyperspectral image classification,'' \emph{IEEE Journal of Selected Topics
  in Applied Earth Observations and Remote Sensing}, vol.~9, no.~2, pp.
  625--639, 2016.

\bibitem{wu2014slow}
C.~Wu, B.~Du, and L.~Zhang, ``Slow feature analysis for change detection in
  multispectral imagery,'' \emph{IEEE Transactions on Geoscience and Remote
  Sensing}, vol.~52, no.~5, pp. 2858--2874, 2014.

\bibitem{lyu2016learning}
H.~Lyu, H.~Lu, and L.~Mou, ``Learning a transferable change rule from a
  recurrent neural network for land cover change detection,'' \emph{Remote
  Sensing}, vol.~8, no.~6, p. 506, 2016.

\bibitem{hu2015transferring}
F.~Hu, G.-S. Xia, J.~Hu, and L.~Zhang, ``Transferring deep convolutional neural
  networks for the scene classification of high-resolution remote sensing
  imagery,'' \emph{Remote Sensing}, vol.~7, no.~11, pp. 14\,680--14\,707, 2015.

\bibitem{olmanson2013airborne}
L.~G. Olmanson, P.~L. Brezonik, and M.~E. Bauer, ``Airborne hyperspectral
  remote sensing to assess spatial distribution of water quality
  characteristics in large rivers: The mississippi river and its tributaries in
  minnesota,'' \emph{Remote Sensing of Environment}, vol. 130, pp. 254--265,
  2013.

\bibitem{moran1997opportunities}
M.~S. Moran, Y.~Inoue, and E.~Barnes, ``Opportunities and limitations for
  image-based remote sensing in precision crop management,'' \emph{Remote
  sensing of Environment}, vol.~61, no.~3, pp. 319--346, 1997.

\bibitem{delalieux2012heathland}
S.~Delalieux, B.~Somers, B.~Haest, T.~Spanhove, J.~V. Borre, and C.~M{\"u}cher,
  ``Heathland conservation status mapping through integration of hyperspectral
  mixture analysis and decision tree classifiers,'' \emph{Remote sensing of
  environment}, vol. 126, pp. 222--231, 2012.

\bibitem{ham2005investigation}
J.~Ham, Y.~Chen, M.~M. Crawford, and J.~Ghosh, ``Investigation of the random
  forest framework for classification of hyperspectral data,'' \emph{IEEE
  Transactions on Geoscience and Remote Sensing}, vol.~43, no.~3, pp. 492--501,
  2005.

\bibitem{melgani2004classification}
F.~Melgani and L.~Bruzzone, ``Classification of hyperspectral remote sensing
  images with support vector machines,'' \emph{IEEE Transactions on geoscience
  and remote sensing}, vol.~42, no.~8, pp. 1778--1790, 2004.

\bibitem{gualtieri2000support}
J.~Gualtieri and S.~Chettri, ``Support vector machines for classification of
  hyperspectral data,'' in \emph{Geoscience and Remote Sensing Symposium, 2000.
  Proceedings. IGARSS 2000. IEEE 2000 International}, vol.~2.\hskip 1em plus
  0.5em minus 0.4em\relax IEEE, 2000, pp. 813--815.

\bibitem{chen2014deep}
Y.~Chen, Z.~Lin, X.~Zhao, G.~Wang, and Y.~Gu, ``Deep learning-based
  classification of hyperspectral data,'' \emph{IEEE Journal of Selected topics
  in applied earth observations and remote sensing}, vol.~7, no.~6, pp.
  2094--2107, 2014.

\bibitem{plaza2009incorporation}
A.~Plaza, J.~Plaza, and G.~Martin, ``Incorporation of spatial constraints into
  spectral mixture analysis of remotely sensed hyperspectral data,'' in
  \emph{Machine Learning for Signal Processing, 2009. MLSP 2009. IEEE
  International Workshop on}.\hskip 1em plus 0.5em minus 0.4em\relax IEEE,
  2009, pp. 1--6.

\bibitem{bioucas2013hyperspectral}
J.~M. Bioucas-Dias, A.~Plaza, G.~Camps-Valls, P.~Scheunders, N.~Nasrabadi, and
  J.~Chanussot, ``Hyperspectral remote sensing data analysis and future
  challenges,'' \emph{IEEE Geoscience and remote sensing magazine}, vol.~1,
  no.~2, pp. 6--36, 2013.

\bibitem{fauvel2008spectral}
M.~Fauvel, J.~A. Benediktsson, J.~Chanussot, and J.~R. Sveinsson, ``Spectral
  and spatial classification of hyperspectral data using svms and morphological
  profiles,'' \emph{IEEE Transactions on Geoscience and Remote Sensing},
  vol.~46, no.~11, pp. 3804--3814, 2008.

\bibitem{li2013spectral}
J.~Li, J.~M. Bioucas-Dias, and A.~Plaza, ``Spectral--spatial classification of
  hyperspectral data using loopy belief propagation and active learning,''
  \emph{IEEE Transactions on Geoscience and Remote Sensing}, vol.~51, no.~2,
  pp. 844--856, 2013.

\bibitem{liu2013spatial}
J.~Liu, Z.~Wu, Z.~Wei, L.~Xiao, and L.~Sun, ``Spatial-spectral kernel sparse
  representation for hyperspectral image classification,'' \emph{IEEE Journal
  of Selected Topics in Applied Earth Observations and Remote Sensing}, vol.~6,
  no.~6, pp. 2462--2471, 2013.

\bibitem{mou2017deep}
L.~Mou, P.~Ghamisi, and X.~X. Zhu, ``Deep recurrent neural networks for
  hyperspectral image classification,'' \emph{IEEE Transactions on Geoscience
  and Remote Sensing}, vol.~55, no.~7, pp. 3639--3655, 2017.

\bibitem{chen2015spectral}
Y.~Chen, X.~Zhao, and X.~Jia, ``Spectral--spatial classification of
  hyperspectral data based on deep belief network,'' \emph{IEEE Journal of
  Selected Topics in Applied Earth Observations and Remote Sensing}, vol.~8,
  no.~6, pp. 2381--2392, 2015.

\bibitem{zhao2016spectral}
W.~Zhao and S.~Du, ``Spectral--spatial feature extraction for hyperspectral
  image classification: A dimension reduction and deep learning approach,''
  \emph{IEEE Transactions on Geoscience and Remote Sensing}, vol.~54, no.~8,
  pp. 4544--4554, 2016.

\bibitem{chen2016deep}
Y.~Chen, H.~Jiang, C.~Li, X.~Jia, and P.~Ghamisi, ``Deep feature extraction and
  classification of hyperspectral images based on convolutional neural
  networks,'' \emph{IEEE Transactions on Geoscience and Remote Sensing},
  vol.~54, no.~10, pp. 6232--6251, 2016.

\bibitem{zhong2018spectral}
Z.~Zhong, J.~Li, Z.~Luo, and M.~Chapman, ``Spectral--spatial residual network
  for hyperspectral image classification: A 3-d deep learning framework,''
  \emph{IEEE Transactions on Geoscience and Remote Sensing}, vol.~56, no.~2,
  pp. 847--858, 2018.

\bibitem{dempster1977maximum}
A.~P. Dempster, N.~M. Laird, and D.~B. Rubin, ``Maximum likelihood from
  incomplete data via the em algorithm,'' \emph{Journal of the royal
  statistical society. Series B (methodological)}, pp. 1--38, 1977.

\bibitem{yarowsky1995unsupervised}
D.~Yarowsky, ``Unsupervised word sense disambiguation rivaling supervised
  methods,'' in \emph{Proceedings of the 33rd annual meeting on Association for
  Computational Linguistics}.\hskip 1em plus 0.5em minus 0.4em\relax
  Association for Computational Linguistics, 1995, pp. 189--196.

\bibitem{blum1998combining}
A.~Blum and T.~Mitchell, ``Combining labeled and unlabeled data with
  co-training,'' in \emph{Proceedings of the eleventh annual conference on
  Computational learning theory}.\hskip 1em plus 0.5em minus 0.4em\relax ACM,
  1998, pp. 92--100.

\bibitem{bie2004convex}
T.~D. Bie and N.~Cristianini, ``Convex methods for transduction,'' in
  \emph{Advances in neural information processing systems}, 2004, pp. 73--80.

\bibitem{blum2001learning}
A.~Blum and S.~Chawla, ``Learning from labeled and unlabeled data using graph
  mincuts,'' 2001.

\bibitem{dopido2013semisupervised}
I.~D{\'o}pido, J.~Li, P.~R. Marpu, A.~Plaza, J.~M.~B. Dias, and J.~A.
  Benediktsson, ``Semisupervised self-learning for hyperspectral image
  classification,'' \emph{IEEE transactions on geoscience and remote sensing},
  vol.~51, no.~7, pp. 4032--4044, 2013.

\bibitem{zhang2014modified}
X.~Zhang, Q.~Song, R.~Liu, W.~Wang, and L.~Jiao, ``Modified co-training with
  spectral and spatial views for semisupervised hyperspectral image
  classification,'' \emph{IEEE Journal of Selected Topics in Applied Earth
  Observations and Remote Sensing}, vol.~7, no.~6, pp. 2044--2055, 2014.

\bibitem{samiappan2015semi}
S.~Samiappan and R.~J. Moorhead, ``Semi-supervised co-training and active
  learning framework for hyperspectral image classification,'' in
  \emph{Geoscience and Remote Sensing Symposium (IGARSS), 2015 IEEE
  International}.\hskip 1em plus 0.5em minus 0.4em\relax IEEE, 2015, pp.
  401--404.

\bibitem{camps2007semi}
G.~Camps-Valls, T.~V.~B. Marsheva, and D.~Zhou, ``Semi-supervised graph-based
  hyperspectral image classification,'' \emph{IEEE Transactions on Geoscience
  and Remote Sensing}, vol.~45, no.~10, pp. 3044--3054, 2007.

\bibitem{ratle2010semisupervised}
F.~Ratle, G.~Camps-Valls, and J.~Weston, ``Semisupervised neural networks for
  efficient hyperspectral image classification,'' \emph{IEEE Transactions on
  Geoscience and Remote Sensing}, vol.~48, no.~5, pp. 2271--2282, 2010.

\bibitem{he2009learning}
H.~He and E.~A. Garcia, ``Learning from imbalanced data,'' \emph{IEEE
  Transactions on knowledge and data engineering}, vol.~21, no.~9, pp.
  1263--1284, 2009.

\bibitem{camps2005kernel}
G.~Camps-Valls and L.~Bruzzone, ``Kernel-based methods for hyperspectral image
  classification,'' \emph{IEEE Transactions on Geoscience and Remote Sensing},
  vol.~43, no.~6, pp. 1351--1362, 2005.

\bibitem{xu2012dcpe}
J.~Xu, H.~He, and H.~Man, ``Dcpe co-training for classification,''
  \emph{Neurocomputing}, vol.~86, pp. 75--85, 2012.

\bibitem{williams1989learning}
R.~J. Williams and D.~Zipser, ``A learning algorithm for continually running
  fully recurrent neural networks,'' \emph{Neural computation}, vol.~1, no.~2,
  pp. 270--280, 1989.

\bibitem{rodriguez1999recurrent}
P.~Rodriguez, J.~Wiles, and J.~L. Elman, ``A recurrent neural network that
  learns to count,'' \emph{Connection Science}, vol.~11, no.~1, pp. 5--40,
  1999.

\bibitem{sundermeyer2015feedforward}
M.~Sundermeyer, H.~Ney, and R.~Schl{\"u}ter, ``From feedforward to recurrent
  lstm neural networks for language modeling,'' \emph{IEEE Transactions on
  Audio, Speech, and Language Processing}, vol.~23, no.~3, pp. 517--529, 2015.

\bibitem{bahdanau2014neural}
D.~Bahdanau, K.~Cho, and Y.~Bengio, ``Neural machine translation by jointly
  learning to align and translate,'' \emph{arXiv preprint arXiv:1409.0473},
  2014.

\bibitem{graves2014towards}
A.~Graves and N.~Jaitly, ``Towards end-to-end speech recognition with recurrent
  neural networks,'' in \emph{International Conference on Machine Learning},
  2014, pp. 1764--1772.

\bibitem{graves2013speech}
A.~Graves, A.-r. Mohamed, and G.~Hinton, ``Speech recognition with deep
  recurrent neural networks,'' in \emph{Acoustics, speech and signal processing
  (icassp), 2013 ieee international conference on}.\hskip 1em plus 0.5em minus
  0.4em\relax IEEE, 2013, pp. 6645--6649.

\bibitem{cho2014properties}
K.~Cho, B.~Van~Merri{\"e}nboer, D.~Bahdanau, and Y.~Bengio, ``On the properties
  of neural machine translation: Encoder-decoder approaches,'' \emph{arXiv
  preprint arXiv:1409.1259}, 2014.

\bibitem{gal2016theoretically}
Y.~Gal and Z.~Ghahramani, ``A theoretically grounded application of dropout in
  recurrent neural networks,'' in \emph{Advances in neural information
  processing systems}, 2016, pp. 1019--1027.

\bibitem{lecun1998gradient}
Y.~LeCun, L.~Bottou, Y.~Bengio, and P.~Haffner, ``Gradient-based learning
  applied to document recognition,'' \emph{Proceedings of the IEEE}, vol.~86,
  no.~11, pp. 2278--2324, 1998.

\bibitem{wang2007analyzing}
W.~Wang and Z.-H. Zhou, ``Analyzing co-training style algorithms,'' in
  \emph{European Conference on Machine Learning}.\hskip 1em plus 0.5em minus
  0.4em\relax Springer, 2007, pp. 454--465.

\bibitem{schuurmans2002metric}
D.~Schuurmans and F.~Southey, ``Metric-based methods for adaptive model
  selection and regularization,'' \emph{Machine Learning}, vol.~48, no. 1-3,
  pp. 51--84, 2002.

\end{thebibliography}

\end{document}